\documentclass[sigconf]{acmart}

\usepackage{algorithm}
\usepackage{algorithmic}
\usepackage{amsmath}

\renewcommand\footnotetextcopyrightpermission[1]{}

\usepackage{newfloat}
\usepackage{listings}
\usepackage{subcaption}
\usepackage{graphicx}
\usepackage{tabularx}
\usepackage{booktabs}
\usepackage{multirow}
\usepackage{amsmath}
\usepackage{makecell} 

\usepackage{booktabs}    
\usepackage{multirow}    
\usepackage{caption}     
\usepackage{booktabs}
\usepackage{tabularx}
\usepackage{colortbl}
\usepackage{multirow}
\usepackage{array}
\usepackage[table]{xcolor}   
\usepackage{colortbl}

\definecolor{oursrow}{RGB}{233,243,253}   
\definecolor{oraclerow}{RGB}{255,243,230} 

\setcopyright{none}

\AtBeginDocument{%
  }

\acmConference[WWW '26]{Proceedings of the ACM Web Conference 2026}{April 13--17, 2026}{Dubai, United Arab Emirates}

\begin{document}

\title{Learning to Route: A Rule-Driven Agent Framework for Hybrid-Source Retrieval-Augmented Generation}

\author{Haoyue Bai}
\email{baihaoyue621@gmail.com}
\affiliation{%
  \institution{Arizona State University}
  \city{Tempe}
  \state{Arizona}
  \country{USA}
}

\author{Haoyu Wang}
\email{haoyu@nec-labs.com}
\authornote{Corresponding Author, the work was accomplished when the first author working as intern in NEC Labs America supervised by the corresponding author}
\affiliation{%
  \institution{NEC Laboratories America}
  \city{Princeton}
  \state{New Jersey}
  \country{USA}
}

\author{Shengyu Chen}
\email{shchen@nec-labs.com}
\affiliation{%
  \institution{NEC Laboratories America}
  \city{Princeton}
  \state{New Jersey}
  \country{USA}
}

\author{Zhengzhang Chen}
\email{zchen@nec-labs.com}
\affiliation{%
  \institution{NEC Laboratories America}
  \city{Princeton}
  \state{New Jersey}
  \country{USA}
}

\author{Lu-An Tang}
\email{ltang@nec-labs.com}
\affiliation{%
  \institution{NEC Laboratories America}
  \city{Princeton}
  \state{New Jersey}
  \country{USA}
}

\author{Wei Cheng}
\email{weicheng@nec-labs.com}
\affiliation{%
  \institution{NEC Laboratories America}
  \city{Princeton}
  \state{New Jersey}
  \country{USA}
}

\author{Yanjie Fu}
\email{yanjie.fu@asu.edu}
\affiliation{%
  \institution{Arizona State University}
  \city{Tempe}
  \state{Arizona}
  \country{USA}
}

\author{Haifeng Chen}
\email{haifeng@nec-labs.com}
\affiliation{%
  \institution{NEC Laboratories America}
  \city{Princeton}
  \state{New Jersey}
  \country{USA}
}

\renewcommand{\shortauthors}{Haoyue Bai et al.}

\begin{abstract}
Large Language Models (LLMs) have shown remarkable performance on general Question Answering (QA), yet they often struggle in domain-specific scenarios where accurate and up-to-date information is required. Retrieval-Augmented Generation (RAG) addresses this limitation by enriching LLMs with external knowledge, but existing systems primarily rely on unstructured documents, while largely overlooking relational databases, which provide precise, timely, and efficiently queryable factual information, serving as indispensable infrastructure in domains such as finance, healthcare, and scientific research. 
Motivated by this gap, we conduct a systematic analysis that reveals three central observations: (i) databases and documents offer complementary strengths across queries, (ii) naively combining both sources introduces noise and cost without consistent accuracy gains, and (iii) selecting the most suitable source for each query is crucial to balance effectiveness and efficiency. We further observe that query types show consistent regularities in their alignment with retrieval paths, suggesting that routing decisions can be effectively guided by systematic rules that capture these patterns. 
Building on these insights, we propose a rule-driven routing framework designed specifically for hybrid-source RAG. 
A routing agent scores candidate augmentation paths based on explicit rules and selects the most suitable one; a rule-making expert agent refines the rules using QA feedback to produce more comprehensive and reliable decision criteria; and a path-level meta-cache reuses past routing decisions for semantically similar queries to reduce latency and cost. 
Experiments on three QA datasets demonstrate that our framework consistently outperforms static strategies and learned routing baselines, achieving higher accuracy while maintaining moderate computational cost.
\end{abstract}




\maketitle


\section{Introduction}

\begin{figure}[ht]
  \centering
  \begin{subfigure}[t]{0.22\textwidth}
    \centering
    \includegraphics[width=\textwidth]{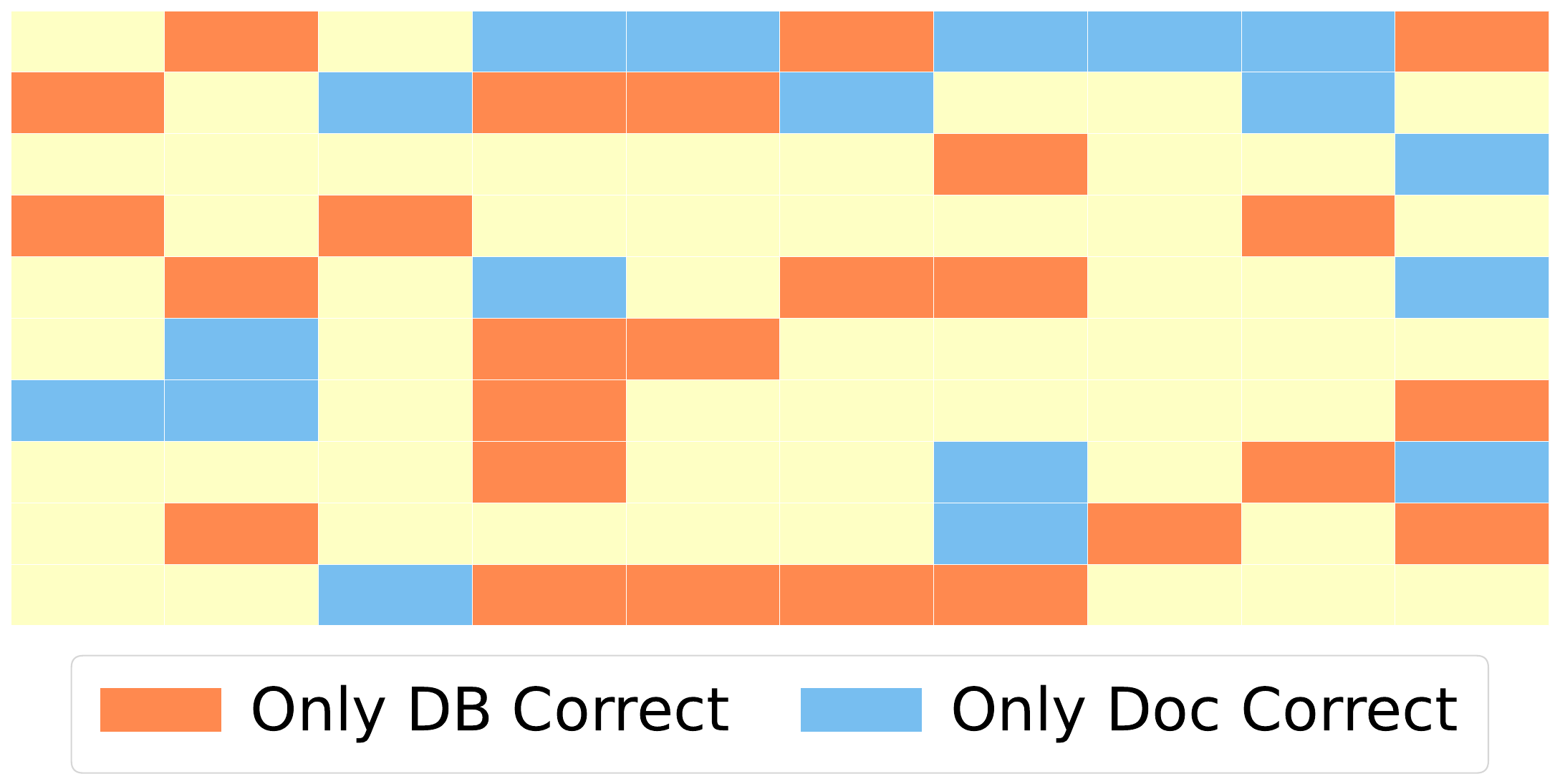}
    \caption{\small Complementarity}
    \label{fig:intro_motivation_1}
  \end{subfigure}
  \hfill
  \begin{subfigure}[t]{0.235\textwidth}
    \centering
    \includegraphics[width=\textwidth]{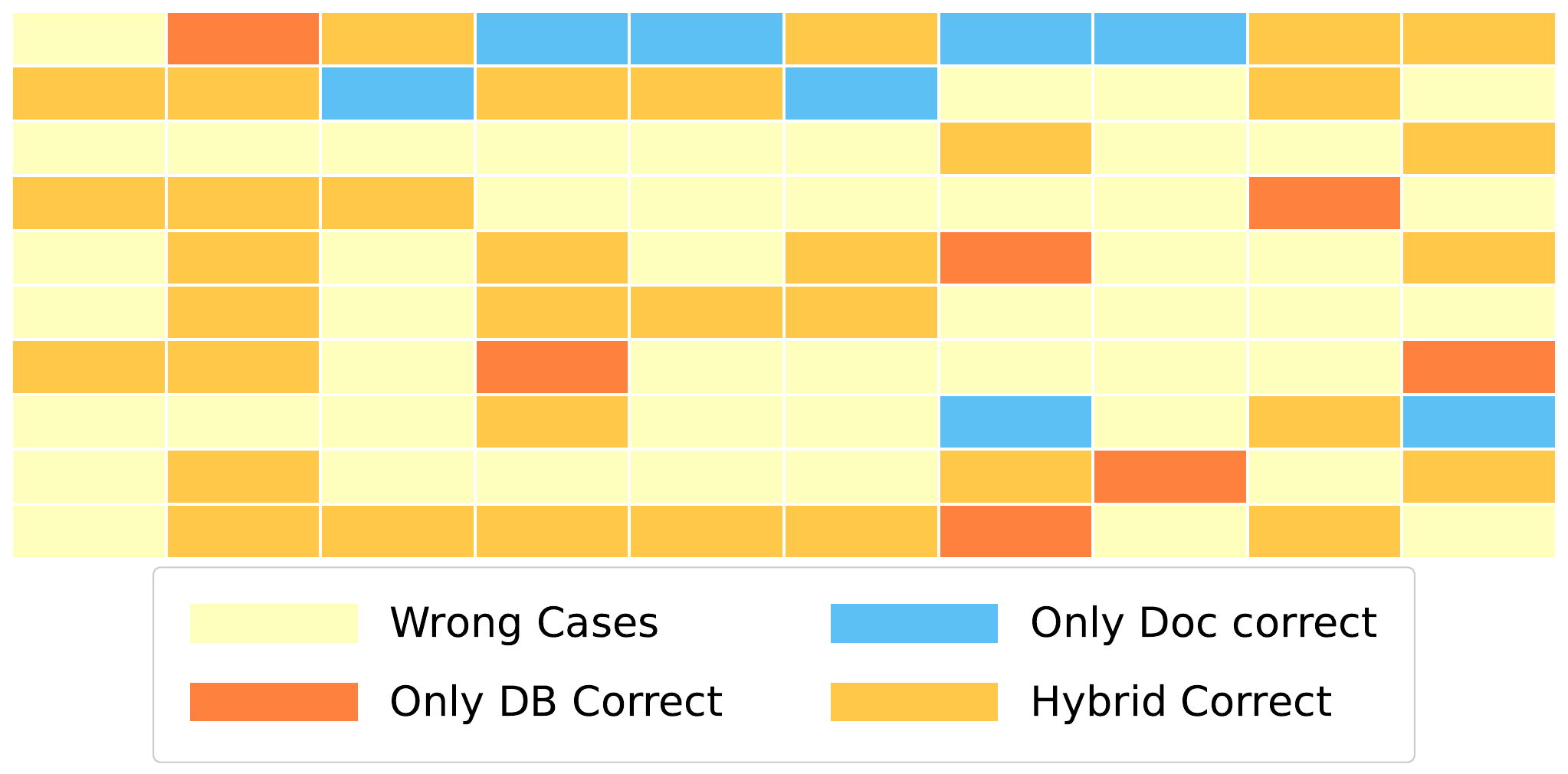}
    \caption{\small Naive Hybridization}
    \label{fig:intro_motivation_2}
  \end{subfigure}


  \begin{subfigure}[t]{0.315\textwidth}
    \centering
    \includegraphics[width=\textwidth]{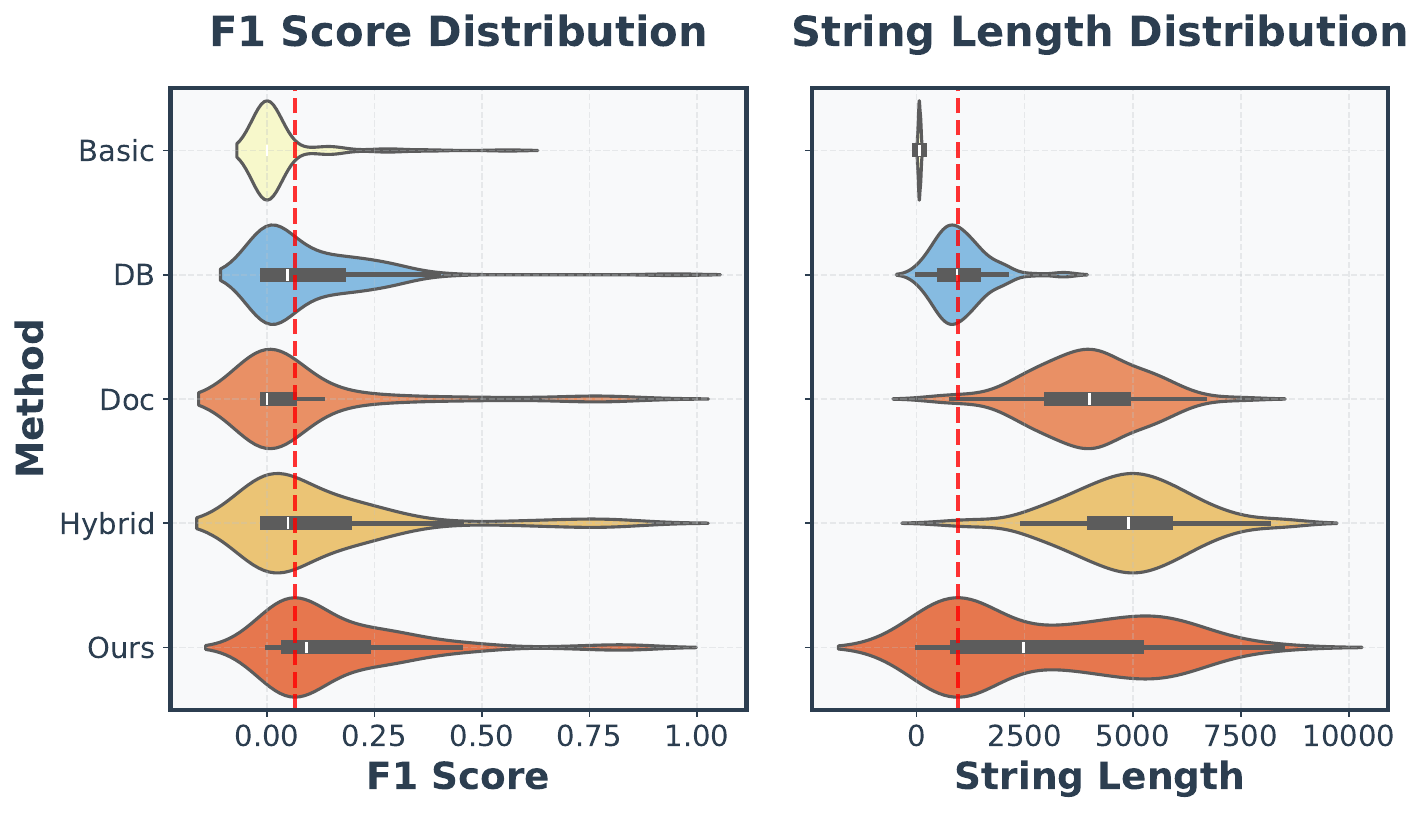}
    \caption{\small Advantages of Routing}
    \label{fig:intro_motivation_3}
  \end{subfigure}
  \hfill
  \begin{subfigure}[t]{0.155\textwidth}
    \centering
    \includegraphics[width=\textwidth]{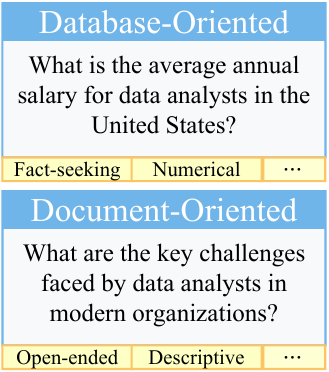}
    \caption{\small Pattern Analysis}
    \label{fig:intro_motivation_4}
  \end{subfigure}

  \caption{\small \textbf{Problem Analysis.} 
  (a) Retrieval Path Effectiveness is Question-Dependent. 
  (b) Excess Context Can Hurt Accuracy. 
  (c) Routing Makes Balance Accuracy and Efficiency. 
  (d) Problem types align with database- vs. document-oriented retrieval.}
  \label{fig:motivation}
\end{figure}


Large language models (LLMs) have achieved remarkable success across a wide range of natural language processing tasks~\cite{shao2026baldro,zhai2026maximizing}, particularly in question answering (QA)~\cite{liu2025debate}. 
Yet, despite their strong capabilities, LLMs still struggle in specialized QA scenarios such as enterprise knowledge access or domain-specific expert consultation, where accurate and up-to-date information is required but may not be captured in the model's parametric memory. 
To address this limitation, Retrieval-Augmented Generation (RAG) has been proposed~\cite{RAG_1,RAG_2,RAG_3}, which enriches LLMs with external knowledge retrieval, thereby improving factual grounding and adaptability.

Existing RAG systems primarily rely on large unstructured corpora, such as Wikipedia or web documents, as their external knowledge base~\cite{DOC_RAG_1,DOC_RAG_2,DOC_RAG_3}. 
Beyond documents, researchers have also explored alternative external sources, such as knowledge graphs for structured entity–relation reasoning~\cite{KG_RAG_1,KG_RAG_2,KG_RAG_3} and semi-structured data like tables or time series~\cite{MM_RAG_1,MM_RAG_2,MM_RAG_3}. 
In contrast, \emph{relational databases} deliver precise, up-to-date, and efficiently queryable factual information~\cite{DB_1,DB_2}. 
They serve as the backbone of critical domains such as finance, healthcare, and scientific research, where reliable access to structured records (e.g., financial transactions, electronic medical records, experimental measurements) is indispensable. 
Unlike unstructured text, databases are designed for accuracy, consistency, and timeliness, ensuring that essential information can be retrieved efficiently and without ambiguity. 
Despite their broad importance, the integration of relational databases into RAG frameworks has received relatively little attention~\cite{DB_RAG_1}.

To better understand the role of relational databases in RAG, we conduct a set of motivating experiments on the TATQA~\cite{tatqa} dataset using GPT-4.1-mini. 
Here, augmentation follows the standard RAG paradigm: the retrieved evidence is appended to the question and provided to the LLM for answer generation, with the only difference being the external knowledge source. 
\textbf{i) Complementarity.} 
We first compare two augmentation paths: one enhanced by relational databases (DB), where query results from relational tables are verbalized into text and fed to the LLM, and the other by unstructured documents (Doc), where relevant passages are retrieved from a document corpus. 
Figure~\ref{fig:intro_motivation_1} visualizes per-question outcomes, where each cell corresponds to one query and the color denotes which augmentation path yields the correct answer. 
The results reveal strong complementarity: many questions can only be answered correctly by DB augmentation, while others require Doc augmentation. 
Neither source dominates across all queries, and relying on a single source leaves substantial coverage gaps. 
\textbf{ii) Naive Hybridization.} 
A natural follow-up is to combine DB and Doc evidence simultaneously by concatenating both sources and passing them to the LLM. 
Figure~\ref{fig:intro_motivation_2} further shows per-question outcomes under this hybrid strategy, where the additional color indicates cases in which both sources are combined. 
We observe that many queries that can be correctly answered by a single path (DB or Doc) may fail under hybrid augmentation. 
Instead, redundant or noisy evidence often distracts the model, leading to incorrect answers, while the token count grows dramatically, increasing inference latency and monetary cost. 
Thus, simply "feeding more" is neither efficient nor reliable. 
\textbf{iii) Necessity of Routing.} 
Finally, we examine the impact of applying a uniform strategy across all queries. 
Figure~\ref{fig:intro_motivation_3} demonstrates that single-source inputs limit accuracy, while hybrid inputs incur high overhead. 
In contrast, an ideal system should allocate each query to its best-suited path, simultaneously improving accuracy and keeping token usage moderate. 
This motivates the need for a routing mechanism that achieves a better balance between \emph{effectiveness} and \emph{efficiency}~\cite{wang-etal-2025-mixllm,shao2024average}.

Therefore, the key challenge is to design an intelligent routing mechanism that, for each query, can dynamically decide whether to rely on DB, Doc, or both. 
While prior studies have examined routing in terms of retrieval complexity~\cite{COMPLEX_ROUTING_1,COMPLEX_ROUTING_2,COMPLEX_ROUTING_3} or retriever choice (e.g., sparse vs. dense)~\cite{BM25,DOC_RAG_1,METHOD_ROUTING_1,METHOD_ROUTING_2}, relatively fewer works have addressed routing across heterogeneous knowledge sources, especially bridging structured relational databases with unstructured documents in a unified manner.
To ground such a mechanism, it is crucial to understand whether queries exhibit systematic patterns that can guide routing decisions.

Beyond aggregate performance, we further observe that queries are not randomly distributed across sources but instead exhibit clear structural regularities. 
In particular, fact-centric or numerical questions are typically better served by database augmentation, whereas open-ended or descriptive queries align more naturally with document retrieval, a pattern consistently observed across multiple datasets (see Figure~\ref{fig:intro_motivation_4} for illustration). 
However, existing learned routers—whether classifier-based or LLM-based—often struggle to stably capture these heterogeneous patterns. 
They require large labeled data to train, behave as black boxes during deployment, and tend to produce uncontrollable routing. 
These limitations motivate a more transparent and rule-grounded approach, where routing decisions are guided by explicit and interpretable rules that encode the observed query–path regularities and capture these understandable patterns. 
Meanwhile, such a rule-driven perspective also naturally resonates with how humans reason and make decisions. 
Cognitive studies show that people rarely exhaust all possible options; rather, they often apply simple "if–else" heuristics to make efficient choices under resource constraints~\cite{human-thinking1,human-thinking2}. 
For example, if it rains, people carry an umbrella; if the weather is clear, they leave it at home. 
These everyday heuristics clearly illustrate how humans rely on interpretable rules to balance accuracy and efficiency without overcomplicating decisions. 
Motivated by this analogy, we design our framework around \emph{rule-driven routing}, where transparent rules encode observed query-path alignments and are incrementally refined through feedback to balance accuracy, efficiency, and interpretability.

Our framework incorporates three key components: 
(i) a \textbf{Rule-Driven Routing Agent}, which avoids the inefficiency of static augmentation by evaluating candidate paths with interpretable rules and selecting the most suitable one; 
(ii) a \textbf{Rule-Making Expert Agent}, which refines these rules using feedback from QA performance, thereby reducing reliance on static, hand-crafted rules and enabling adaptation to different datasets; and
(iii) a \textbf{Path-Level Meta-Cache}, which further accelerates inference by reusing routing decisions for repeated or semantically similar queries, eliminating unnecessary agent calls. 
Together, these components enable adaptive, interpretable, and efficient integration of structured and unstructured knowledge sources. 
Our contributions are summarized as follows:
\begin{itemize}
    \item We conduct analysis of routing between relational databases and document corpora in RAG, highlighting their complementary strengths through motivating experiments.  
    \item We introduce a rule-driven routing framework that integrates a routing agent, a rule-making expert agent, and a path-level meta-cache, aiming to balance accuracy, interpretability, and efficiency.
    \item  We evaluate the proposed framework on three QA benchmarks, showing that it achieves consistent improvements over static strategies and alternative routing baselines, while maintaining moderate cost.
\end{itemize}

\section{Related Work}
\subsection{Hybrid Knowledge Sources for RAG}
Retrieval-Augmented Generation (RAG) is an influential framework that enhances the capabilities of large language models (LLMs) by enabling access to external knowledge sources, thereby improving their performance on tasks that require domain-specific information~\cite{RAG_1,RAG_2,RAG_3}.
Unlike traditional language models, which generate text based solely on internal parameters, RAG models incorporate a retrieval component that fetches relevant content from external data sources and integrates it into the generation process.
Traditional RAG systems primarily utilize large-scale unstructured corpora, such as Wikipedia and web documents, as their external knowledge base~\cite{DOC_RAG_1,DOC_RAG_2,DOC_RAG_3}. 
Building on this foundation, more recent work has expanded the spectrum of retrieval sources to further enhance RAG's capabilities.
Some studies integrate knowledge graphs, which provide structured representations of entities and relationships. Knowledge graphs enable more context-aware and entity-centric retrieval, supporting complex information needs and facilitating multi-hop reasoning~\cite{KG_RAG_1,KG_RAG_2,KG_RAG_3}.
Other approaches incorporate multi-modal data—such as images, videos, and tables—enabling RAG systems to synthesize information across modalities and improving their robustness on heterogeneous, complex queries~\cite{MM_RAG_1,MM_RAG_2,MM_RAG_3}.
In contrast, relational databases store highly structured and precise factual data, supporting efficient updates and fast queries~\cite{DB_1,DB_2}. 
Despite their widespread use in many enterprise and scientific applications, the integration of relational databases as knowledge sources in RAG has received relatively little attention~\cite{DB_RAG_1}.
There is a lack of systematic analysis and methodical approaches for combining the complementary strengths of relational databases and unstructured documents within RAG frameworks.
Addressing this gap, our work aims to explore and propose strategies for the joint utilization of these two data sources.

\subsection{Routing Mechanisms in RAG}

Routing mechanisms have received increasing attention in RAG systems, enabling dynamic and intelligent decision-making about how each query is processed. 
Unlike fixed pipelines, routing-enhanced RAG architectures employ dedicated modules that analyze the characteristics of incoming queries and adaptively select the most appropriate retrieval and reasoning pathways, thereby improving both efficiency and response quality.
A fundamental role of routing in RAG is to determine the appropriate complexity of the retrieval pipeline for each query. While some questions can be resolved through simple, single-hop retrieval, others require more complex, multi-hop reasoning or iterative retrieval procedures~\cite{COMPLEX_ROUTING_1,COMPLEX_ROUTING_2,COMPLEX_ROUTING_3}. Adaptive routing allows the system to dynamically decide whether external retrieval is needed or if the language model's parametric knowledge suffices, effectively balancing computational cost and answer accuracy.
Another crucial dimension of routing concerns the choice of retrieval methods. Early RAG systems often used simple rule-based or classifier-driven approaches, sending queries with strong lexical overlap to sparse retrievers such as BM25~\cite{BM25}, while directing semantically complex or ambiguous queries to dense retrievers~\cite{DOC_RAG_1}. More recent advances have introduced lightweight classifiers, neural models, and multi-agent frameworks that can select or combine multiple retrieval strategies, sometimes explicitly modeling the interplay between external retrievals and the language model's internal knowledge~\cite{METHOD_ROUTING_1,METHOD_ROUTING_2}.
Beyond retrieval strategy, routing also governs data source selection. In real-world scenarios, the knowledge necessary to answer a query may reside in document collections, knowledge bases, or other sources. Modern RAG systems increasingly deploy specialized routers—such as domain routers or dynamism routers—to intelligently map queries to the most relevant data sources and to orchestrate hybrid retrieval workflows that capitalize on the complementary advantages of each data type~\cite{DATA_ROUTING_1,DATA_ROUTING_2,DATA_ROUTING_3}.
In this work, we focus on dynamic routing and the integration of structured relational databases with unstructured documents in RAG systems. By recognizing and leveraging the distinct strengths of these sources, we propose a routing mechanism that adaptively selects the knowledge source.

\section{Preliminaries}

\noindent \textbf{Question-Answering.}
Given a natural language question $q$, the most basic question-answering (QA) approach with modern LLMs is to directly generate an answer $a$ based solely on the model's internal (parametric) knowledge, i.e., $a = \mathrm{LLM}(q)$. While extremely efficient, such answers are often limited by the LLM’s overall coverage, timeliness, and factual reliability.

\noindent \textbf{Document-Enhanced QA.}
We then introduce document-enhanced QA, where external unstructured documents are incorporated into the answer generation process. Specifically, for each question $q$, a retrieval module (e.g., BM25~\cite{BM25} or dense retriever~\cite{DOC_RAG_1}) searches a pre-built index of document passages, and the top-$k$ retrieved segments $\mathcal{D}_q = \{\text{doc}_1, \ldots, \text{doc}_k\}$ are provided as additional context to the LLM: $a = \mathrm{LLM}(q, \mathcal{D}_q)$. This mechanism allows the LLM to access more domain-specific information during inference.
\begin{figure*}[ht]
\centering
\includegraphics[width = 1\textwidth]{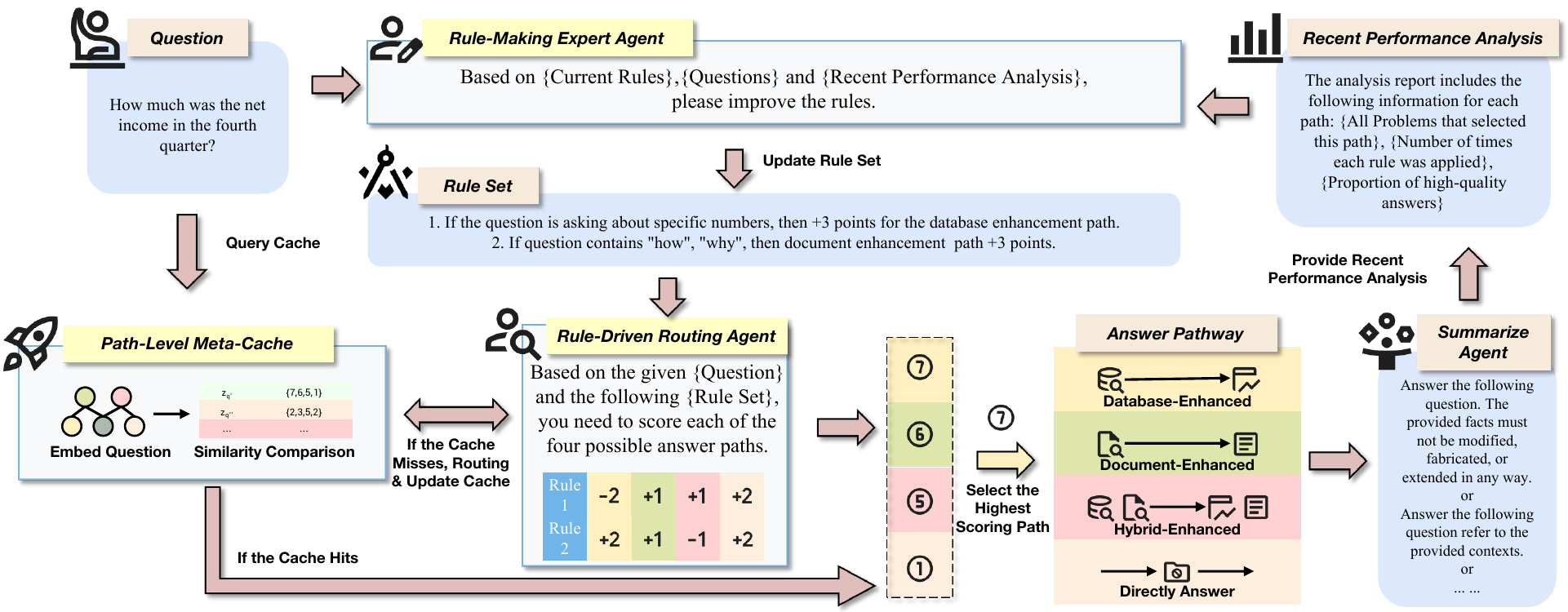}
\caption{\small 
Overall workflow of our rule-driven routing framework. 
At inference, each question first checks the \textbf{Path-Level Meta-Cache}; if a hit occurs, the cached path scores are reused, otherwise the query is passed to the \textbf{Rule-Driven Routing Agent}, which scores candidate paths based on explicit rules and updates the cache. 
The highest-scoring path (Database, Document, Hybrid, or Direct Answer) is then selected to provide evidence to the \textbf{Summarize Agent} for answer generation. 
During training, question–answer performance is collected and analyzed, and the \textbf{Rule-Making Expert Agent} refines the rule set, enabling subsequent routing decisions to adaptively improve.
}
\label{fig:overall}
\end{figure*}
\noindent \textbf{Database-Enhanced QA.}  
We further consider database-enhanced QA, which leverages structured relational data to provide precise, time-sensitive, and verifiable answers.  
For a given question $q$, the system first retrieves relevant tables by matching $q$ against a pre-built index of table metadata, including schemas, high-frequency values, and LLM-generated table descriptions.  
Once candidate tables are identified, factual records $\mathcal{F}_q$ are extracted from the database either through advanced text-to-SQL generation methods~\cite{Text2SQL1,Text2SQL2} or via efficient keyword-based filtering.  
The resulting records serve as immutable evidence and are incorporated into the final prompt: $ a = \mathrm{LLM}(q, \mathcal{F}_q)$. In our implementation, we explicitly instruct the LLM that retrieved database facts are not to be modified or reinterpreted, ensuring that the generated answers remain precise.


\noindent \textbf{Hybrid-Enhanced QA.}
We further enable hybrid-enhanced QA. Here, for each query, both the top-ranked document passages $\mathcal{D}_q$ and the extracted database facts $\mathcal{F}_q$ are retrieved and fused as context for the LLM: $a = \mathrm{LLM}(q, \mathcal{D}_q, \mathcal{F}_q)$. This hybridization aims to maximize answer coverage, informativeness, and trustworthiness. However, aggregating more information inevitably increases inference time and computational cost, and in some cases, introducing excessive context may even hurt answer quality.

\noindent \textbf{Routing Mechanisms.}
The complementary nature of document and database augmentation presents a new challenge: for each question, how to dynamically select the most effective enhancement pathway to balance answer quality, response latency, and computational cost. We formalize this as a query routing problem: given a question $q$, a routing agent selects an augmentation path $p_q \in \mathcal{P} = {\texttt{Doc}, \texttt{DB}, \texttt{Hybrid}, \texttt{LLM}}$ according to a policy that optimizes a utility objective combining quality, latency, and cost. The selected pathway determines which information is retrieved and used to construct the prompt for the downstream LLM, which generates the final answer. This design allows the system to flexibly adapt to each query’s needs, leveraging the complementary strengths of both data sources while managing efficiency.


\section{Methodology}
\subsection{Overall Framework}
Relational databases are well-suited for precise and fast retrieval, especially when queries target clearly structured facts or up-to-date information. 
However, their coverage is often limited and they struggle with vague, open-ended, or semantic questions. 
In contrast, document corpora offer broader coverage and are more robust to semantic or fuzzy queries, but they are prone to hallucination and their longer context  can increase both latency and cost.
To address these trade-offs and enable flexible, high-quality retrieval-augmented QA, we propose a rule-driven routing framework (Figure ~\ref{fig:overall}). For each incoming query $q$, it integrates (i) a Rule-Driven Routing Agent for path scoring and selection, (ii) a Rule-Making Expert Agent for iterative, data-driven rule refinement, (iii) a Path-Level Meta-Cache for efficient and safe decision reuse.

\subsection{Rule-Driven Routing Agent}
Question intents in such scenarios often follow template-like patterns, which makes them particularly suitable for rule-based routing (e.g., questions starting with "how" typically call for document augmentation). 
Rule-driven routing is training-free, inherently interpretable, and readily supports domain adaptation and expert intervention. 
By leveraging language models as the routing agent, the system can operationalize not only simple keyword patterns but also more semantic and context-dependent rules—significantly enhancing flexibility and robustness compared to rigid rule engines.  
Concretely, the agent takes a query $q$, evaluates it against a set of rules $\mathcal{R}$, assigns scores to candidate augmentation paths, and selects the highest-scoring path for downstream QA. 
This design operationalizes human-like heuristics in a principled way, making the routing process flexible, robust, and easy to interpret while retaining the ability to capture nuanced query intents.

\noindent \textbf{Expert-Initialized Rules.}
Our routing design begins with a set of interpretable rules $\mathcal{R}$, initialized by domain experts.  
Each rule assigns additive scores to candidate augmentation paths based on recognizable query patterns.  
Formally, we denote by $S_p(q)$ the score assigned to path $p$ for query $q$, where higher values indicate stronger preference for selecting that path.  
A subset of illustrative rules (complete rules are provided in Appendix \ref{appendix:rule}) is as follows:  
\[\begin{aligned}
\text{Rule 1: } & \text{If query requests numbers, } S_{\texttt{DB}}(q) {+}{=} 3. \\
\text{Rule 2: } & \text{If query contains "how," "why", } S_{\texttt{Doc}}(q) {+}{=} 3. \\
\text{Rule 3: } & \text{If query seeks definitions, } S_{\texttt{LLM}}(q) {+}{=} 3. \\
\text{Rule 4: } & \text{If query seeks fact with explanation, } S_{\texttt{Hybrid}}(q) {+}{=} 3. \\
\end{aligned}\]
These rules directly shape routing scores in a transparent manner, ensuring that the reasoning process is both interpretable and auditable. 
Much like human natural decision-making, such "if–else" heuristics provide simple yet effective guidance.

\noindent \textbf{Routing Mechanism.}  
We introduce a rule-driven routing agent to $\mathrm{A}_{\mathrm{ROUTING}}$ score each candidate path $p \in \mathcal{P}$ for a given query $q$. The routing policy selects the path with the maximum score:
\begin{equation}
p_q = \arg\max_{p \in \mathcal{P}} S_p(q),
\end{equation}
where the score is produced by the agent conditioned on both the query and rule set:
\begin{equation}
S_p(q) = \mathrm{A}_{\mathrm{ROUTING}}(q, p, \mathcal{R}).
\end{equation}
When multiple paths tie for the highest score, a predefined priority order is applied to break ties, ensuring stability and transparency in routing decisions.  

\noindent \textbf{Illustrative Example.}  
Consider the query: \textit{"How much was the net income in the fourth quarter?"}.  
According to Rule~1, the request for a numerical fact linked to a specific time yields an increased score for the \texttt{DB} path. Competing options such as \texttt{Doc} or \texttt{LLM} receive lower scores, as the query is fact-centric rather than descriptive or definitional. Consequently, the agent routes the query to the database path, ensuring both accuracy and efficiency.  
In practice, the routing agent applies a richer and more rigorous set of rules, which can capture overlapping conditions and subtle query intents.

\subsection{Rule-Making Expert Agent}
While expert-initialized rules provide a transparent starting point, they may not fully capture the characteristics of a target dataset or benchmark. 
To reduce reliance on static hand-crafted rules, we introduce a \textbf{rule-making expert agent} $\mathcal{A}_{\mathrm{RULE}}$, which refines the rule set using available training or validation data before deployment. 
The workflow is as follows: after a batch of queries is processed, the system records both routing decisions and QA outcomes. 
The expert agent then receives the current rule set $\mathcal{R}$ together with a structured summary of these results, and produces an updated rule set $\mathcal{R}^{(t+1)}$ that better aligns routing with utility objectives (e.g., accuracy, efficiency, or their trade-off). 
This refinement can be carried out offline as a pre-deployment step, ensuring that the rules are better adapted to the specific domain, while keeping the online inference process lightweight and efficient. 
Much like how humans refine simple heuristics through prior experience before applying them in practice, this step improves the practicality and reliability of rule-driven routing without incurring additional runtime cost.

\noindent \textbf{Rule Performance Diagnostics.}  
After processing a batch of queries, the system generates a diagnostic report to support empirical rule refinement. 
This report contains four components:  
(i) the queries themselves,  
(ii) the current rule set $\mathcal{R}$,  
(iii) path-level statistics, such as the selection frequency and accuracy of each augmentation path, and  
(iv) rule-level statistics, including the number of times each rule was triggered and the accuracy achieved when triggered.  
Together, these diagnostics provide a principled basis for evaluating which rules remain effective, which may require adjustment, and where new rules might be beneficial.

\noindent \textbf{Textual Gradient for Rule Updates.}  
Given the diagnostics $\mathcal{M}^{(t)}$, the expert agent produces refinements to the rule set by interpreting performance feedback in natural language, effectively serving as a textual gradient.  
This gradient identifies which rules should be strengthened, weakened, or redefined, and proposes modifications accordingly. 
Formally, rule evolution is defined as:
\begin{equation}
\mathcal{R}^{(t+1)} = \mathcal{A}_{\mathrm{RULE}}(\mathcal{R}^{(t)}, \mathcal{M}^{(t)}),
\end{equation}
where $\mathcal{A}_{\mathrm{RULE}}$ denotes the expert agent and $\mathcal{M}^{(t)}$ represents the diagnostics at the update step $t$. 
By iteratively applying this process, the system refines its routing policy to reflect utility requirements while keeping the online inference process unaffected.

\begin{table*}[ht]
\centering
\renewcommand{\arraystretch}{1.2} 
\small
\setlength{\tabcolsep}{4pt}
\caption{Performance Comparison across Models and Methods (F1 / Accuracy)}
\begin{tabularx}{\textwidth}{@{}l l *{6}{>{\centering\arraybackslash}X}@{}}
\specialrule{2pt}{1pt}{1pt}
\multicolumn{2}{c}{\multirow{2}{*}{Method}} & \multicolumn{2}{c}{TATQA} & \multicolumn{2}{c}{FinQA} & \multicolumn{2}{c}{WikiQA} \\
\cmidrule(lr){3-4} \cmidrule(lr){5-6} \cmidrule(lr){7-8}
& & F1 & Acc & F1 & Acc & F1 & Acc \\
\specialrule{1pt}{2pt}{1pt}
\multicolumn{2}{c}{Basic QA} & 0.0350 & 0.050 & 0.0008 & 0.010 & 0.0432 & 0.138 \\
\multicolumn{2}{c}{Doc} & 0.0608 & 0.100 & 0.0022 & 0.028 & 0.0679 & 0.236 \\
\multicolumn{2}{c}{DB} & 0.0530 & 0.150 & 0.0032 & 0.040 & 0.0513 & 0.188 \\
\multicolumn{2}{c}{Hybrid} & 0.0872 & 0.190 & 0.0016 & 0.030 & 0.0899 & 0.280 \\
\specialrule{1pt}{2pt}{1pt}
\multicolumn{2}{c}{Rule-Based} & 0.0656 & 0.170 & 0.0032 & 0.040 & 0.0868 & 0.186 \\
\multicolumn{2}{c}{Adaptive-RAG} & 0.0345 & 0.180 & 0.0012 & 0.008 & 0.0940 & 0.158 \\
\specialrule{1pt}{2pt}{1pt}
\multirow{5}{*}{LLaMA-3} 
& Agent-Based & 0.0576 $\pm$ 0.0017 & 0.130 $\pm$ 0.0051 & 0.0017 $\pm$ 0.0002 & 0.014 $\pm$ 0.0004 & 0.0911 $\pm$ 0.0028 & 0.152 $\pm$ 0.0049 \\
& Rule Agent & 0.0747 $\pm$ 0.0112 & 0.176 $\pm$ 0.0050 & 0.0036 $\pm$ 0.0001 & 0.040 $\pm$ 0.0019 & 0.0812 $\pm$ 0.0032 & 0.234 $\pm$ 0.0087 \\
& Score Agent & 0.0737 $\pm$ 0.0091 & 0.188 $\pm$ 0.0035 & \underline{0.0037} $\pm$ 0.0013 & 0.038 $\pm$ 0.0156 & 0.0935 $\pm$ 0.0031 & 0.260 $\pm$ 0.0093 \\
\rowcolor{blue!5}
& Ours & \textbf{0.0936} $\pm$ 0.0082 & \textbf{0.212} $\pm$ 0.0182 & 0.0033 $\pm$ 0.0002 & \underline{0.040} $\pm$ 0.0056 & \textbf{0.0987} $\pm$ 0.0039 & \textbf{0.288} $\pm$ 0.0074 \\
\rowcolor{blue!5}
& Ours-c & \underline{0.0900} $\pm$ 0.0031 & \underline{0.194} $\pm$ 0.0025 & \textbf{0.0042} $\pm$ 0.0005 & \textbf{0.046} $\pm$ 0.0053 & \underline{0.0961} $\pm$ 0.0021 & \underline{0.282} $\pm$ 0.0151 \\
\specialrule{1pt}{2pt}{1pt}
\multirow{5}{*}{Qwen2.5} 
& Agent-Based & 0.0800 $\pm$ 0.0026 & 0.188 $\pm$ 0.0057 & 0.0037 $\pm$ 0.0001 & 0.046 $\pm$ 0.0021 & \underline{0.0884} $\pm$ 0.0033 & 0.258 $\pm$ 0.0069 \\
& Rule Agent & 0.0790 $\pm$ 0.0023 & 0.182 $\pm$ 0.0051 & 0.0037 $\pm$ 0.0002 & 0.042 $\pm$ 0.0018 & 0.0815 $\pm$ 0.0025 & 0.240 $\pm$ 0.0074 \\
& Score Agent & 0.0751 $\pm$ 0.0124 & 0.180 $\pm$ 0.0149 & 0.0038 $\pm$ 0.0003 & 0.042 $\pm$ 0.0020 & 0.0855 $\pm$ 0.0026 & 0.240 $\pm$ 0.0063 \\
\rowcolor{blue!5}
& Ours & \textbf{0.0978} $\pm$ 0.0117 & \textbf{0.220} $\pm$ 0.0185 & \underline{0.0045} $\pm$ 0.0005 & \underline{0.050} $\pm$ 0.0043 & \textbf{0.0970} $\pm$ 0.0042 & \textbf{0.302} $\pm$ 0.0092 \\
\rowcolor{blue!5}
& Ours-c & \underline{0.0944} $\pm$ 0.0012 & \underline{0.218} $\pm$ 0.0011 & \textbf{0.0047} $\pm$ 0.0002 & \textbf{0.054} $\pm$ 0.0021 & 0.0882 $\pm$ 0.0024 & \underline{0.290} $\pm$ 0.0064 \\
\specialrule{1pt}{2pt}{1pt}
\multirow{5}{*}{GPT-4o} 
& Agent-Based & 0.0693 $\pm$ 0.0007 & 0.1380 $\pm$ 0.0009 & 0.0040 $\pm$ 0.0003 & 0.0360 $\pm$ 0.0028 & 0.0911 $\pm$ 0.0016 & 0.2020 $\pm$ 0.0041 \\
& Rule Agent & 0.0800 $\pm$ 0.0026 & 0.1820 $\pm$ 0.0074 & 0.0039 $\pm$ 0.0001 & 0.0420 $\pm$ 0.0019 & 0.0868 $\pm$ 0.0014 & 0.2280 $\pm$ 0.0060 \\
& Score Agent & 0.0796 $\pm$ 0.0014 & 0.1960 $\pm$ 0.0068 & 0.0037 $\pm$ 0.0001 & 0.0430 $\pm$ 0.0009 & \underline{0.0919} $\pm$ 0.0010 & 0.2480 $\pm$ 0.0109 \\
\rowcolor{blue!5}
& Ours & \textbf{0.0991} $\pm$ 0.0048 & \textbf{0.2200} $\pm$ 0.0123 & \textbf{0.0045} $\pm$ 0.0003 & \textbf{0.0480} $\pm$ 0.0021 & \textbf{0.0930} $\pm$ 0.0019 & \textbf{0.2620} $\pm$ 0.0116 \\
\rowcolor{blue!5}
& Ours-c & \underline{0.0952} $\pm$ 0.0041 & \underline{0.2120} $\pm$ 0.0098 & \underline{0.0042} $\pm$ 0.0004 & \underline{0.0460} $\pm$ 0.0017 & 0.0915 $\pm$ 0.0022 & \underline{0.2500} $\pm$ 0.0103 \\
\specialrule{1pt}{2pt}{1pt}
\multirow{5}{*}{GPT-4.1} 
& Agent-Based & 0.0602 $\pm$ 0.0028 & 0.158 $\pm$ 0.0087 & 0.0034 $\pm$ 0.0004 & 0.042 $\pm$ 0.0020 & 0.0905 $\pm$ 0.0025 & 0.214 $\pm$ 0.0056 \\
& Rule Agent & 0.0709 $\pm$ 0.0024 & \underline{0.180} $\pm$ 0.0053 & 0.0033 $\pm$ 0.0005 & 0.040 $\pm$ 0.0019 & 0.0913 $\pm$ 0.0031 & 0.228 $\pm$ 0.0062 \\
& Score Agent & 0.0726 $\pm$ 0.0029 & 0.178 $\pm$ 0.0044 & 0.0038 $\pm$ 0.0008 & 0.042 $\pm$ 0.0020 & \underline{0.0915} $\pm$ 0.0027 & 0.208 $\pm$ 0.0054 \\
\rowcolor{blue!5}
& Ours & \textbf{0.0779} $\pm$ 0.0037 & \textbf{0.184} $\pm$ 0.0151 & \textbf{0.0042} $\pm$ 0.0012 & \textbf{0.046} $\pm$ 0.0038 & \textbf{0.0954} $\pm$ 0.0054 & \underline{0.242} $\pm$ 0.0082 \\
\rowcolor{blue!5}
& Ours-c & \underline{0.0754} $\pm$ 0.0051 & 0.178 $\pm$ 0.0294 & \underline{0.0040} $\pm$ 0.0007 & \underline{0.042} $\pm$ 0.0024 & \underline{0.0915} $\pm$ 0.0036 & \textbf{0.246} $\pm$ 0.0071 \\
\specialrule{1pt}{2pt}{1pt}
\multicolumn{2}{c}{Oracle} & 0.1020 & 0.264 & 0.0046 & 0.062 & 0.1050 & 0.398 \\
\specialrule{2pt}{1pt}{1pt}

\end{tabularx}
\label{tab:performance-comparison}
\end{table*}

\subsection{Path-Level Meta-Cache}
While our rule-driven routing agent substantially reduces the inefficiency of naive hybrid augmentation, it still requires an additional agent call for rule evaluation.
This extra step introduces some overhead, yet the cost is minor compared to the savings achieved by avoiding redundant hybrid augmentation.
Nevertheless, in large-scale deployments, even small latencies can accumulate, motivating further optimization.
Since many real-world queries are repeated or semantically similar, caching is a natural choice to further reduce overhead.
However, in dynamic scenarios such as frequently updated relational databases, traditional answer-level caches may become unreliable.
For example, caching answers to queries like "What is the revenue growth over the past 7 days?" can easily lead to stale or incorrect results as underlying data change.
To overcome this limitation, we propose a \textbf{path-level meta-cache} that operates at the routing decision level rather than the answer level.

Our meta-cache stores routing decisions—specifically, the embedding representation of each query, the scores for all candidate augmentation paths, and the selected path—for previous queries. 
This structure enables fast and generalized decision-making for repeated or semantically similar questions while fully preserving the reliability of factual responses. 
The meta-cache leverages latent-space (embedding) similarity for flexible retrieval, making it particularly well-suited for dynamic or time-sensitive data environments.
This design yields three main advantages: (i) reduced computation and latency, by bypassing LLM-based routing for many queries; (ii) semantic generalization, by supporting approximate reuse via embedding similarity; and (iii) robust applicability, as it avoids the pitfalls of stale answer reuse in evolving databases.
Formally, let each query $q$ be mapped into a latent representation (embedding) $z_q = \phi(q)$, where $\phi(\cdot)$ denotes the embedding function (e.g., a sentence transformer~\cite{Sentence_trans}). The cache is then
\begin{equation}
\mathcal{C} = \left\{(z_{q_j}, S_{\texttt{Doc}}(q_j),  S_{\texttt{DB}}(q_j), S_{\texttt{Hybrid}}(q_j), S_{\texttt{LLM}}(q_j))\right\},
\end{equation}
where $S_{p}(q_j)$ is the score for path $p \in \mathcal{P} = \{\texttt{Doc}, \texttt{DB}, \texttt{Hybrid}, \texttt{LLM}\}$ for previous query $q_j$. For a new query $q'$, with embedding $z_{q'}$, the system first checks if there exists a cached $z_{q_j}$ such that
\begin{equation}
\text{sim}(z_{q_j}, z_{q'}) \geq \tau,
\end{equation}
where $\tau$ is a threshold used to balance the probability and the accuracy of hitting.
If so, the routing decision and path scores from the matched entry are directly reused; otherwise, the system falls back to the full rule-driven routing process and updates the cache by adding the current query.

\section{Experiment}

\subsection{Experiment Setting}
\noindent\textbf{Datasets.}
We selected three datasets that include both structured database support and unstructured document support for experiments. TATQA~\cite{tatqa} and FINQA~\cite{finqa} are financial report datasets. WIKIQA is a general knowledge question answering dataset. For WIKIQA, We use data from WTQ~\cite{wtq} as the structured dataset and content from Wikipedia~\cite{DOC_RAG_1} as the unstructured text. The test questions are constructed from SQuAD~\cite{squad} and WTQ~\cite{wtq}.

\noindent\textbf{Baselines.}  
We compare our method against two categories of baselines.  
First, we consider \emph{non-dynamic pathways} that adopt a fixed augmentation strategy: directly answering without retrieval (Basic), document-enhanced retrieval (Doc), database-enhanced retrieval (DB), and hybrid-enhanced retrieval (Hybrid).  
Second, we evaluate \emph{dynamic routing strategies}, which adaptively select augmentation paths for each query.  
Among them, lightweight approaches include a keyword-driven rule-based router that maps queries to paths using handcrafted rules~\cite{rule-based}.  
We also include \emph{Adaptive-RAG}~\cite{COMPLEX_ROUTING_1}, which trains a classifier to predict the most suitable path for a given query.  
Finally, we compare against several LLM-enabled routing methods: (i) \emph{Agent-Based}, where the LLM directly chooses a path from the query~\cite{DB_RAG_1}; (ii) \emph{Rule Agent}, where the LLM selects paths based on explicit rules~\cite{rule-agent}; and (iii) \emph{Score Agent}, where the LLM scores rules and selects the path with the highest score~\cite{score-agent}.  

\begin{figure*}[ht]
  \centering
  \begin{subfigure}[t]{0.25\textwidth}
    \centering
    \includegraphics[width=\textwidth]{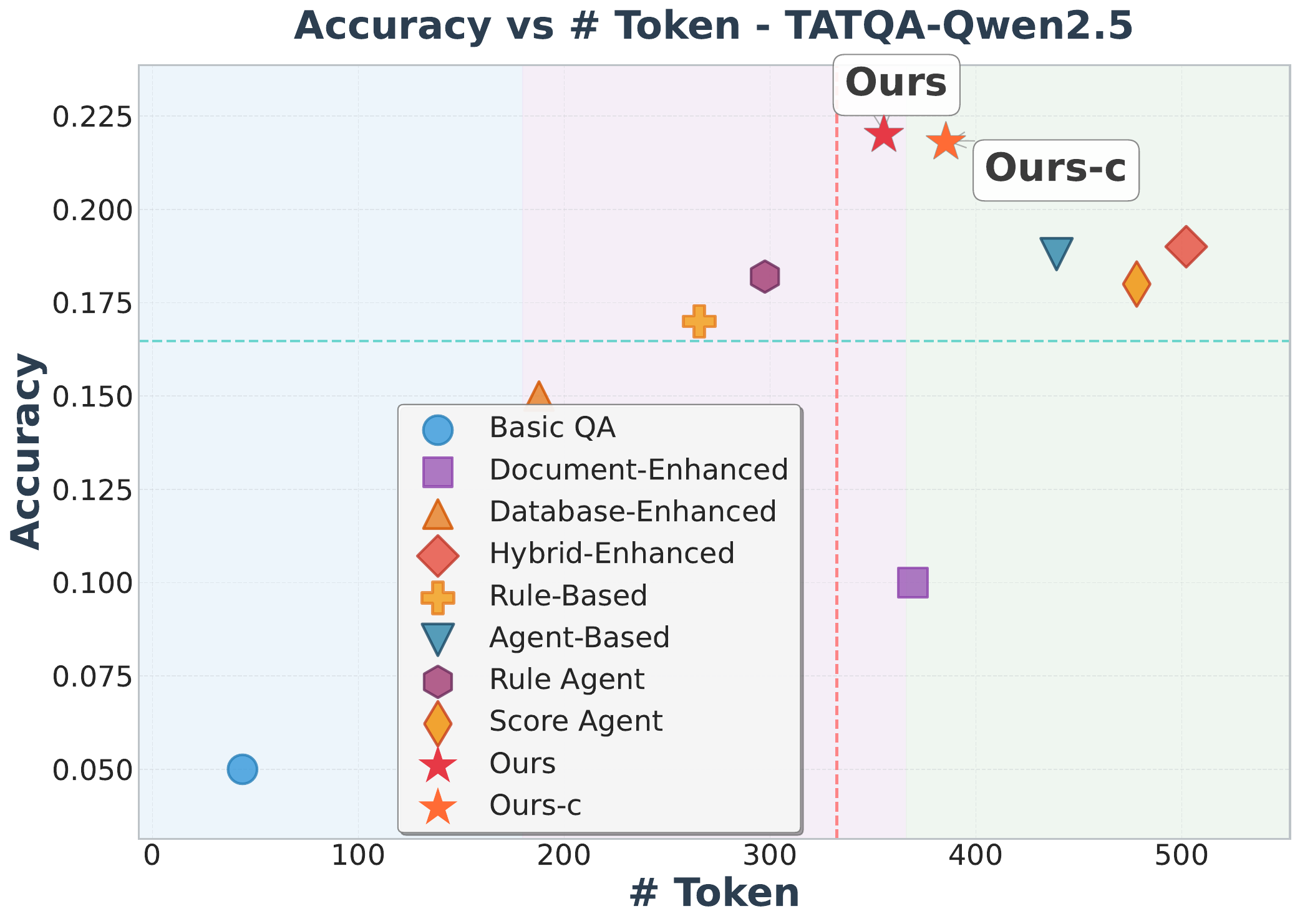}
    \caption{Accuracy vs \#Token - TATQA}
    \label{fig:vs1}
  \end{subfigure}%
  \hfill
  \begin{subfigure}[t]{0.25\textwidth}
    \centering
    \includegraphics[width=\textwidth]{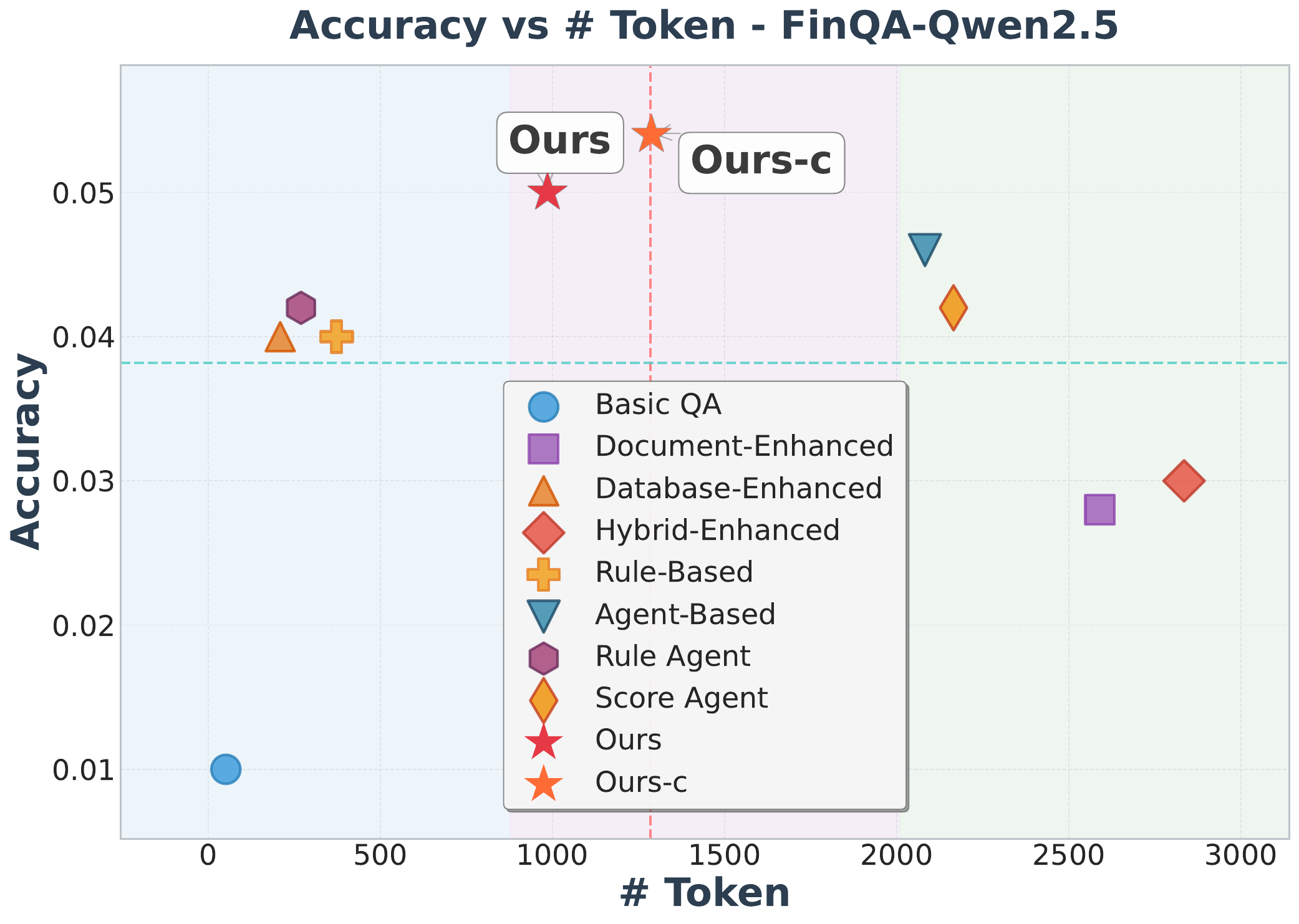}
    \caption{Accuracy vs \#Token - FinQA}
    \label{fig:vs2}
  \end{subfigure}%
  \hfill
  \begin{subfigure}[t]{0.235\textwidth}
    \centering
    \includegraphics[width=\textwidth]{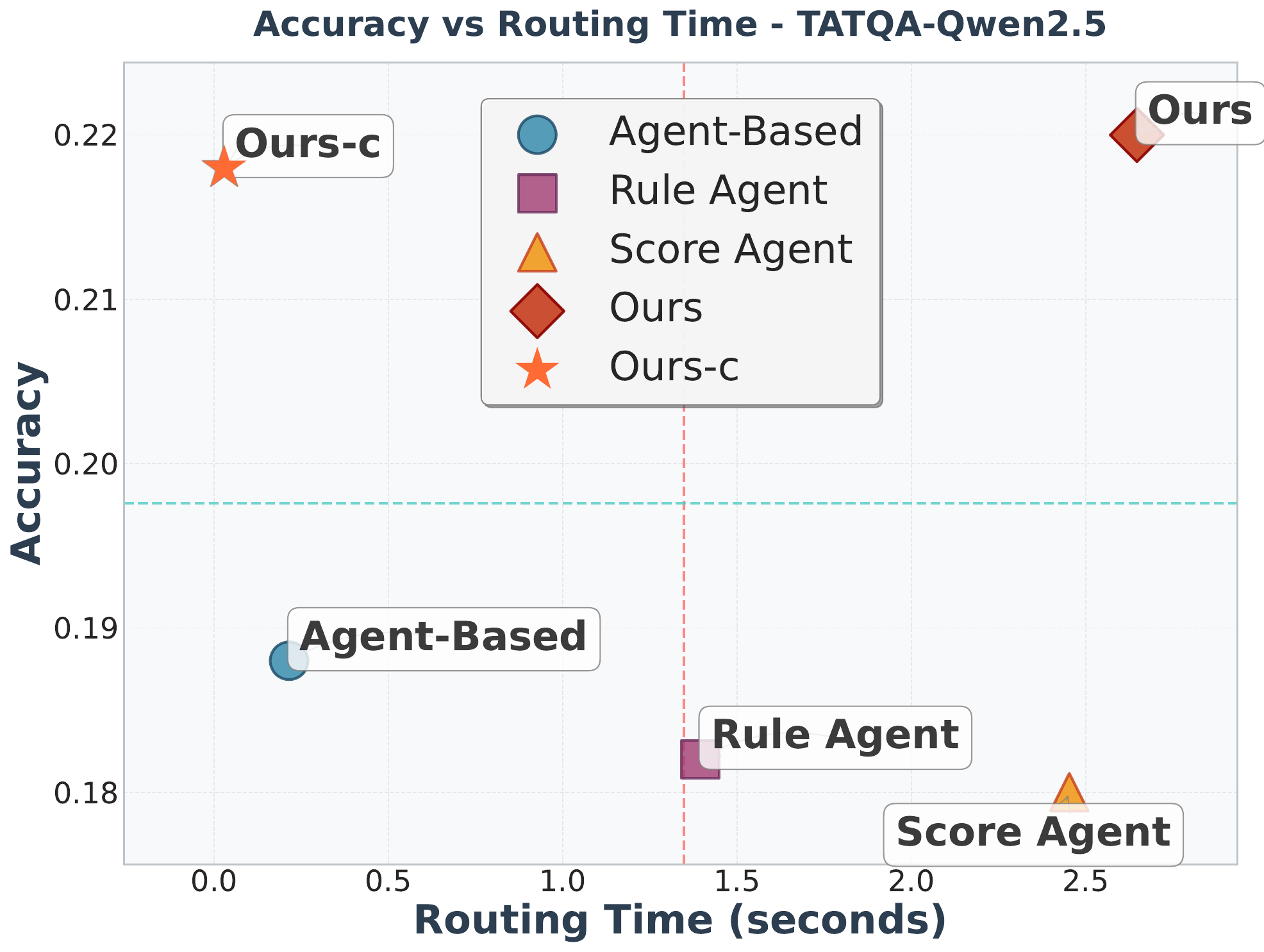}
    \caption{Accuracy vs Time - TATQA}
    \label{fig:vs3}
  \end{subfigure}%
  \hfill
  \begin{subfigure}[t]{0.235\textwidth}
    \centering
    \includegraphics[width=\textwidth]{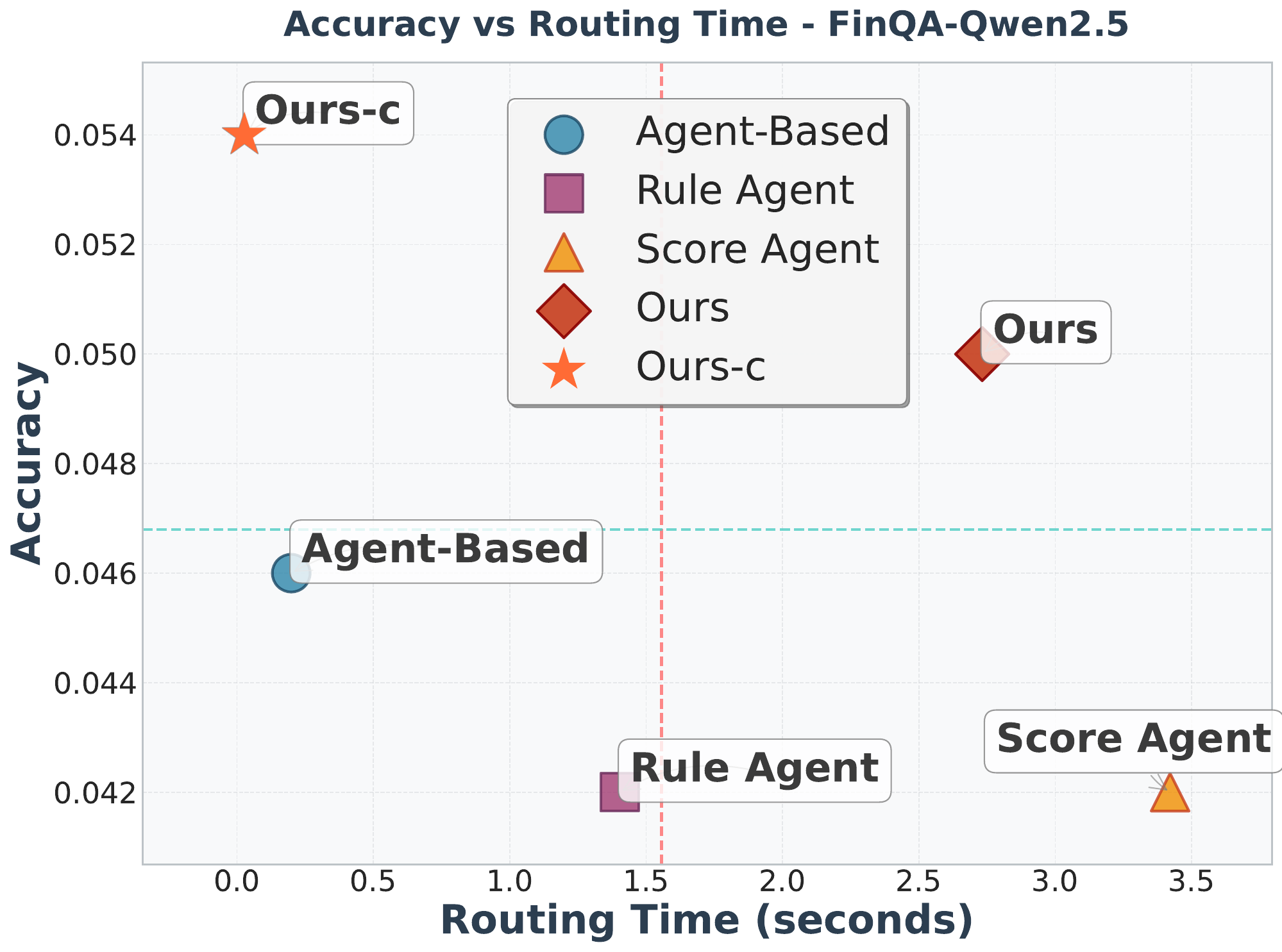}
    \caption{Accuracy vs Time - FinQA}
    \label{fig:vs4}
  \end{subfigure}

  \caption{\small Accuracy and Efficiency Analysis}
  \label{fig:motivation}
\end{figure*}

\noindent\textbf{Implementation.}
We adopt a consistent retrieval setup across all pathways to ensure comparability. Specifically, we use BM25~\cite{BM25}, a classical term-based sparse retrieval model, to retrieve candidate documents or table metadata depending on the selected pathway. For answer generation, we employ GPT-4.1-mini to produce responses under each augmentation pathway. To guarantee fairness, the routing policy is restricted to selecting among these pre-computed answers rather than re-generating them, so that differences stem solely from routing choices rather than generation variability.
Following prior work~\cite{500,COMPLEX_ROUTING_1}, we randomly sample 500 queries for evaluation. In settings that require training data (e.g., for rule refinement), we allocate an additional set of 100 queries as the training set. For both the routing agent and the rule-making expert agent, we experiment with three backbone LLMs: LLaMA-3-8B-Instruct (abbreviated as LLaMA-3)~\cite{llama3modelcard}, Qwen2.5-14B-Instruct (Qwen2.5)~\cite{qwen2.5}, GPT-4o-mini (GPT-4o) and GPT-4.1-mini (GPT-4.1). All experiments are repeated multiple times with different random seeds, and we report the mean and standard deviation across runs.

\subsection{Performance Comparison with Existing Methods}

This experiment aims to evaluate whether dynamic routing can provide more accurate QA performance compared to single-path and static hybrid strategies. 
The results in Table~\ref{tab:performance-comparison} highlight two key observations. 
First, routing-based strategies generally outperform single augmentation paths, demonstrating the advantage of adaptively selecting the most suitable information source. 
Interestingly, in several cases the routing baselines even surpass the hybrid strategy, indicating that blindly concatenating multiple sources can introduce noise, while effective routing helps identify the correct path and avoid unnecessary context.
Second, our proposed framework consistently outperforms all routing baselines across datasets and model backbones. 
For instance, on TATQA with LLaMA-3, our approach improves accuracy from 0.188 (Score Agent) to \textbf{0.212}, and on WikiQA with Qwen2.5 it achieves a new state-of-the-art accuracy of \textbf{0.302}, well above the best competing baseline (0.260). 
These results confirm that our method delivers more accurate and stable performance, validating the effectiveness of the designed routing mechanism.



\subsection{Accuracy vs. Token Efficiency}

This experiment aims to examine the trade-off between model accuracy and computational cost. 
Since the number of tokens directly correlates with both inference latency and monetary expense, we adopt it as a proxy measure of efficiency. 
Figure~\ref{fig:vs1} and ~\ref{fig:vs2} show the accuracy-token relationship on TATQA and FinQA under the Qwen2.5 backbone.
On TATQA, our methods (Ours and Ours-c) achieve accuracies above 0.21 while keeping token counts around 300, whereas the hybrid strategy consumes over 400 tokens for noticeably lower accuracy. 
Similarly, on FinQA, both Ours and Ours-c dominate the upper-left region of the plot, offering clear accuracy advantages without the excessive token usage exhibited by document-heavy strategies.  
We observe that our proposed method achieves the highest accuracy across both datasets, while maintaining only moderate token usage. 
In contrast, static hybrid strategies often incur large token overhead without clear accuracy gains, and some routing baselines sacrifice performance to reduce cost. 
These results demonstrate that our framework not only delivers superior QA accuracy, but also sustains a favorable balance between effectiveness and efficiency.

\subsection{Accuracy vs. Routing Time}

This experiment investigates the fundamental trade-off between routing efficiency and overall QA performance. 
Routing time is a particularly critical factor in real-world systems, as it directly reflects the latency introduced before answer generation. 
Figure~\ref{fig:vs3} and Figure~\ref{fig:vs4} present the relationship between accuracy and routing time on TATQA and FinQA using the Qwen2.5 backbone.  
On TATQA, our method Ours-c attains the highest accuracy while requiring almost negligible routing time, clearly outperforming conventional rule-based or score-based agents that consume over 2 seconds on average with noticeably lower accuracy. 
Similarly, on FinQA, Ours-c again dominates the upper-left corner, delivering consistently superior accuracy with minimal routing latency, while alternatives such as Score Agent and Rule Agent remain both slower and less accurate.  
These results confirm that our framework achieves a highly favorable balance by providing higher accuracy at substantially lower routing cost, highlighting its distinctive advantage for deployment in latency-sensitive applications.

\begin{figure*}[htbp]
  \centering
  \subfloat[\centering 
  \label{fig:RUT}
  \small{Effect of Rule Update on TATQA}]{
    \includegraphics[width=0.19\textwidth]{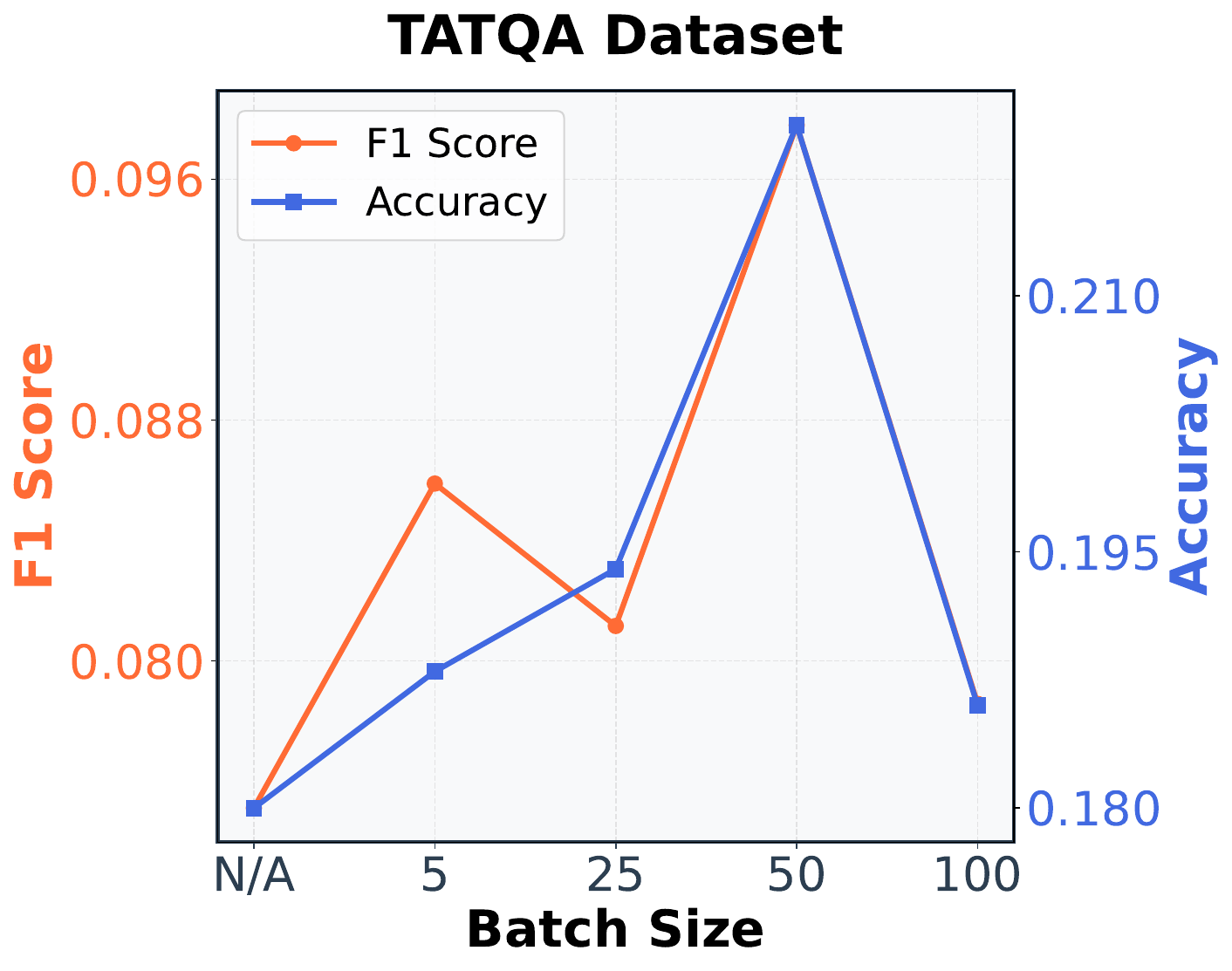}
  }
    \subfloat[\centering 
    \label{fig:RUF}
    \small{Effect of Rule Update on FINQA}]{
    \includegraphics[width=0.19\textwidth]{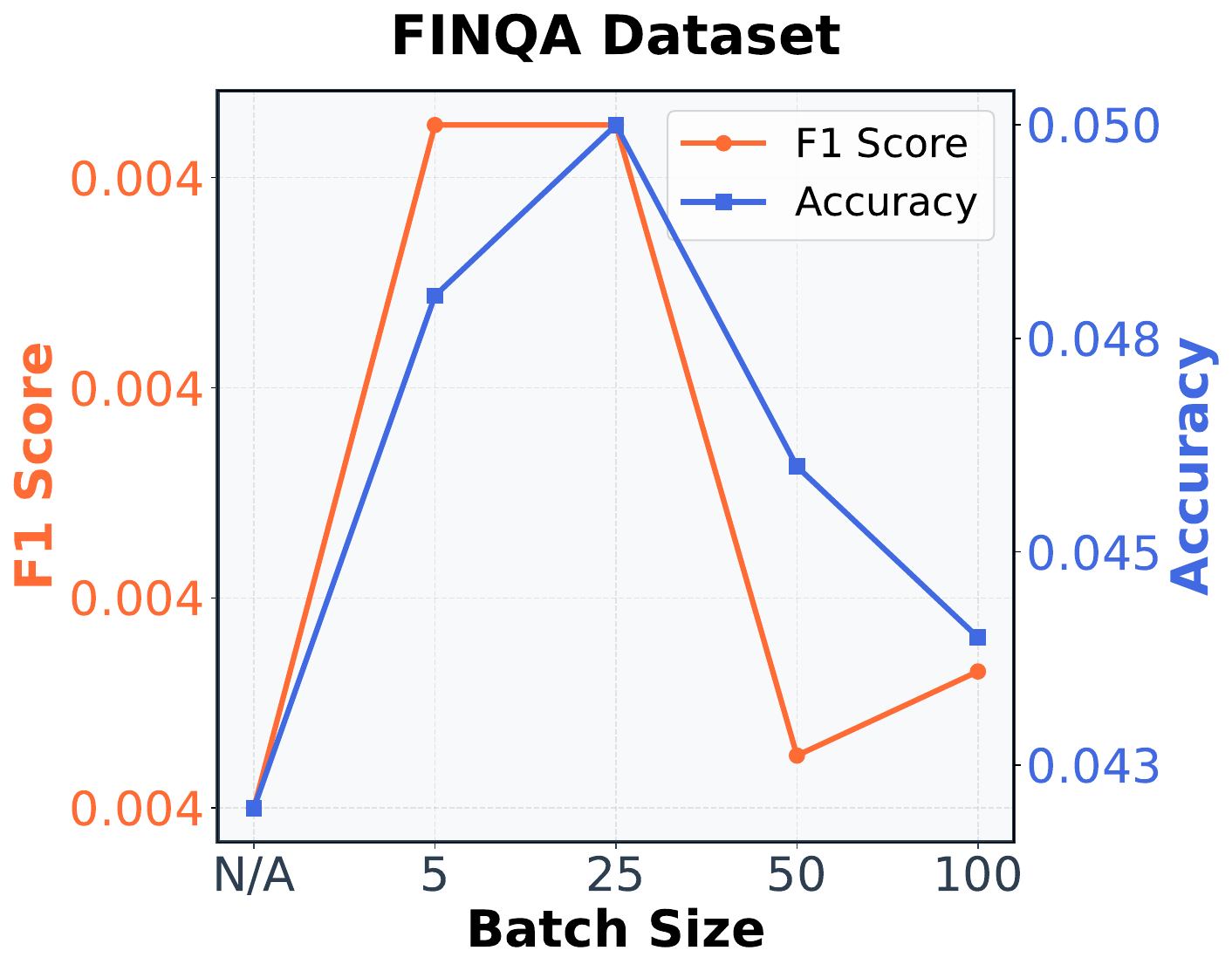}
  }
    \subfloat[\centering 
    \label{fig:PDT}
    \small{Path Detail on TATQA}]{
    \includegraphics[width=0.205\textwidth]{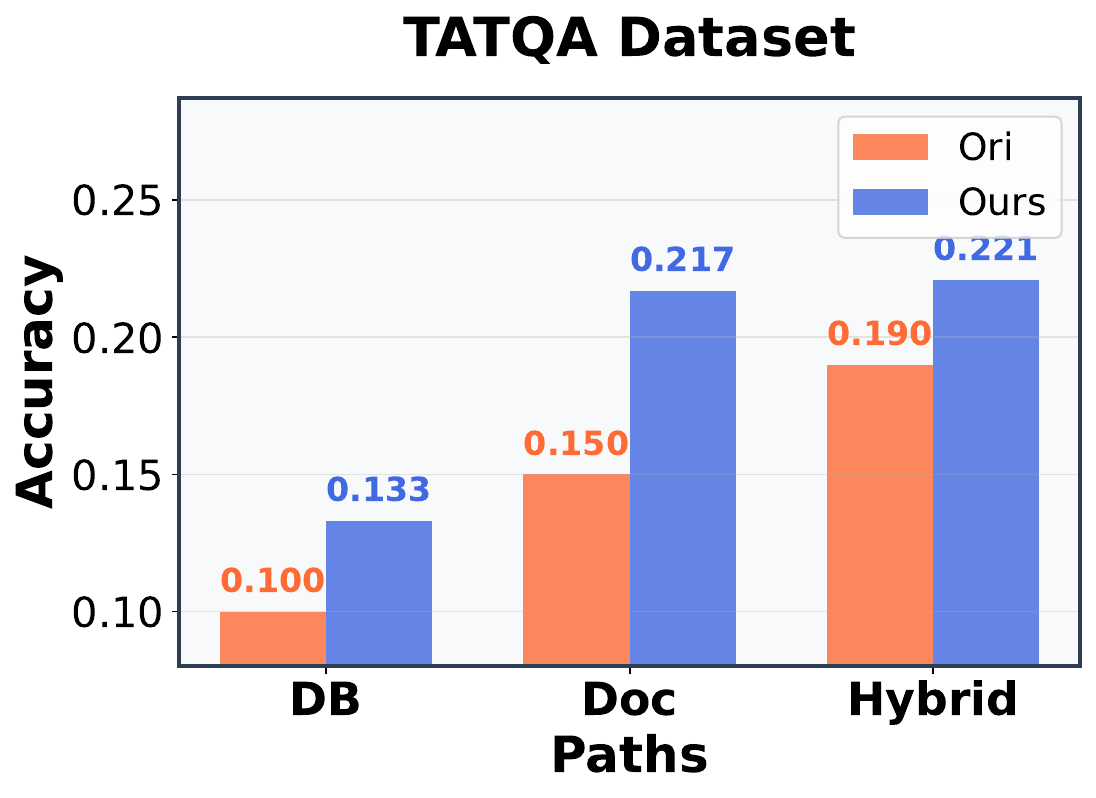}
  }
  \subfloat[\centering 
    \label{fig:PDF}
    \small{Path Detail on FINQA}]{
    \includegraphics[width=0.205\textwidth]{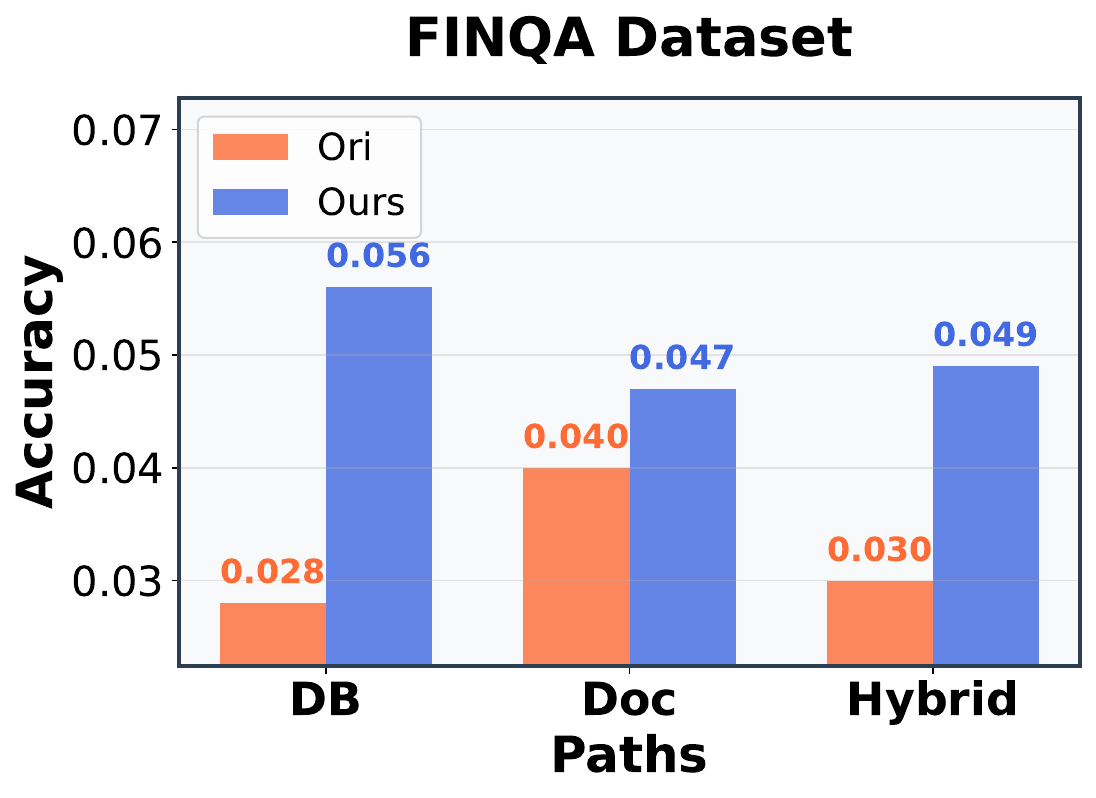}
  }
  \subfloat[\centering 
  \label{fig:PDW}
  \small{Path Detail on WIKIQA}]{
    \includegraphics[width=0.205\textwidth]{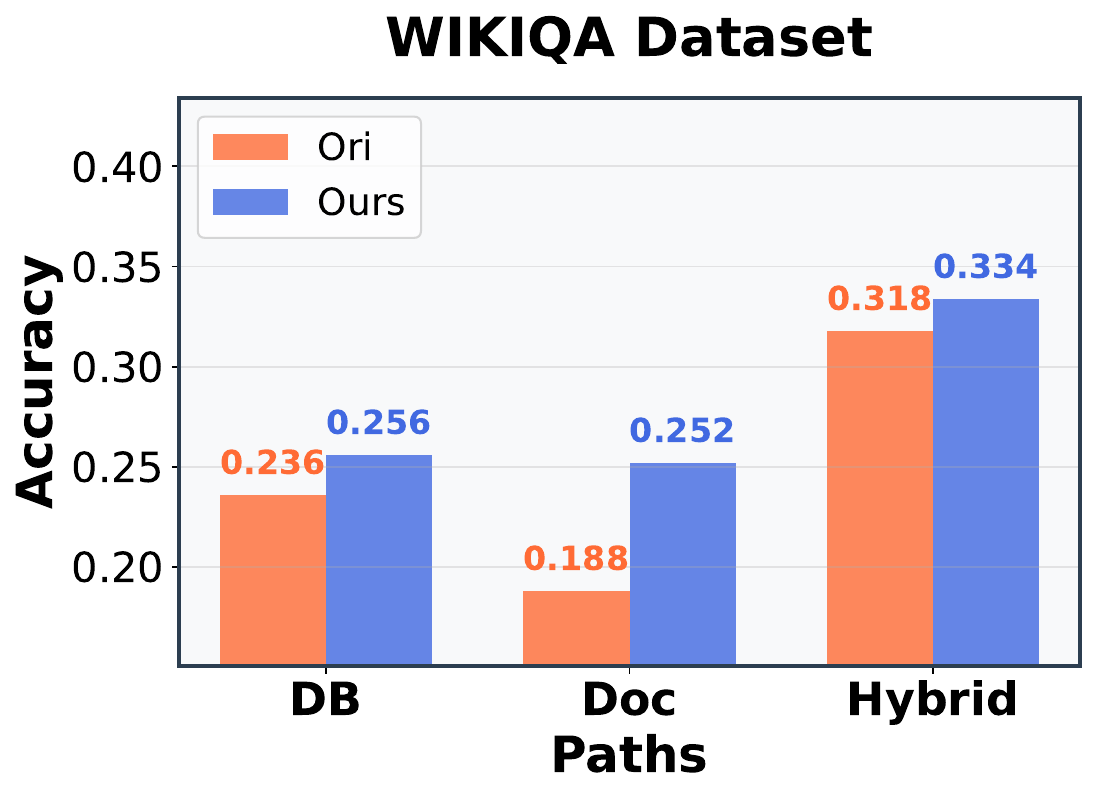}
  }
\caption{\small{Investigation of Proposed Method}}\label{fig:Investigation}
\end{figure*}

\subsection{Effect of Rule Update Frequency}

This experiment investigates the impact of frequency on the effectiveness of rule updates performed by the expert agent. 
Since rule updates refine the routing policy based on accumulated query–answer feedback, the frequency of updates determines how quickly the system adapts to data distribution.
Figure~\ref{fig:RUT} and Figure~\ref{fig:RUF} present the results on TATQA and FinQA. 
We first observe that even a single update, corresponding to batch size $100$, already improves performance compared with the no-update case. 
Moreover, moderate batch sizes, such as $25$ or $50$, further enhance both accuracy and F1 score, showing that more frequent updates help the routing agent align with empirical outcomes. 
For example, on TATQA, the F1 score increases steadily from $0.080$ without updates to over $0.096$ when batch size is set to $50$. 
On FinQA, accuracy peaks when the batch size is $25$, indicating that timely adaptation yields the most effective routing policy. 
These results confirm that incorporating rule updates is consistently beneficial, and that an appropriate update frequency enables the framework to achieve further improvements by better capturing evolving query characteristics.

\subsection{Analysis of Path Utilization}

This experiment examines how routing affects the effectiveness of individual augmentation paths. 
In Figure~\ref{fig:PDT}–\ref{fig:PDW}, the bar plots compare two settings: the accuracy obtained when all queries are forced through a single augmentation path (orange), versus the accuracy of the same path when it is selectively chosen by our routing mechanism (blue). 
This comparison isolates the effect of routing on path quality, independent of the overall distribution of queries across paths.
We observe that across all datasets, selectively chosen paths consistently achieve higher accuracy than their single-path baselines. 
For instance, on TATQA as well as FinQA and WikiQA, the DB, Doc, and Hybrid paths all show clear accuracy gains when routing assigns only the most suitable queries to them. 
These results demonstrate that routing not only improves overall system accuracy but also enhances each individual augmentation strategy by avoiding inefficient one-size-fits-all usage and instead leveraging path complementarity in a targeted and adaptive manner.

\begin{figure}[htbp]
  \centering
  \subfloat[
  \centering 
  \label{fig:PSR_tatqa}
  \small{Path Selection on TATQA}]{
    \includegraphics[width=0.22\textwidth]{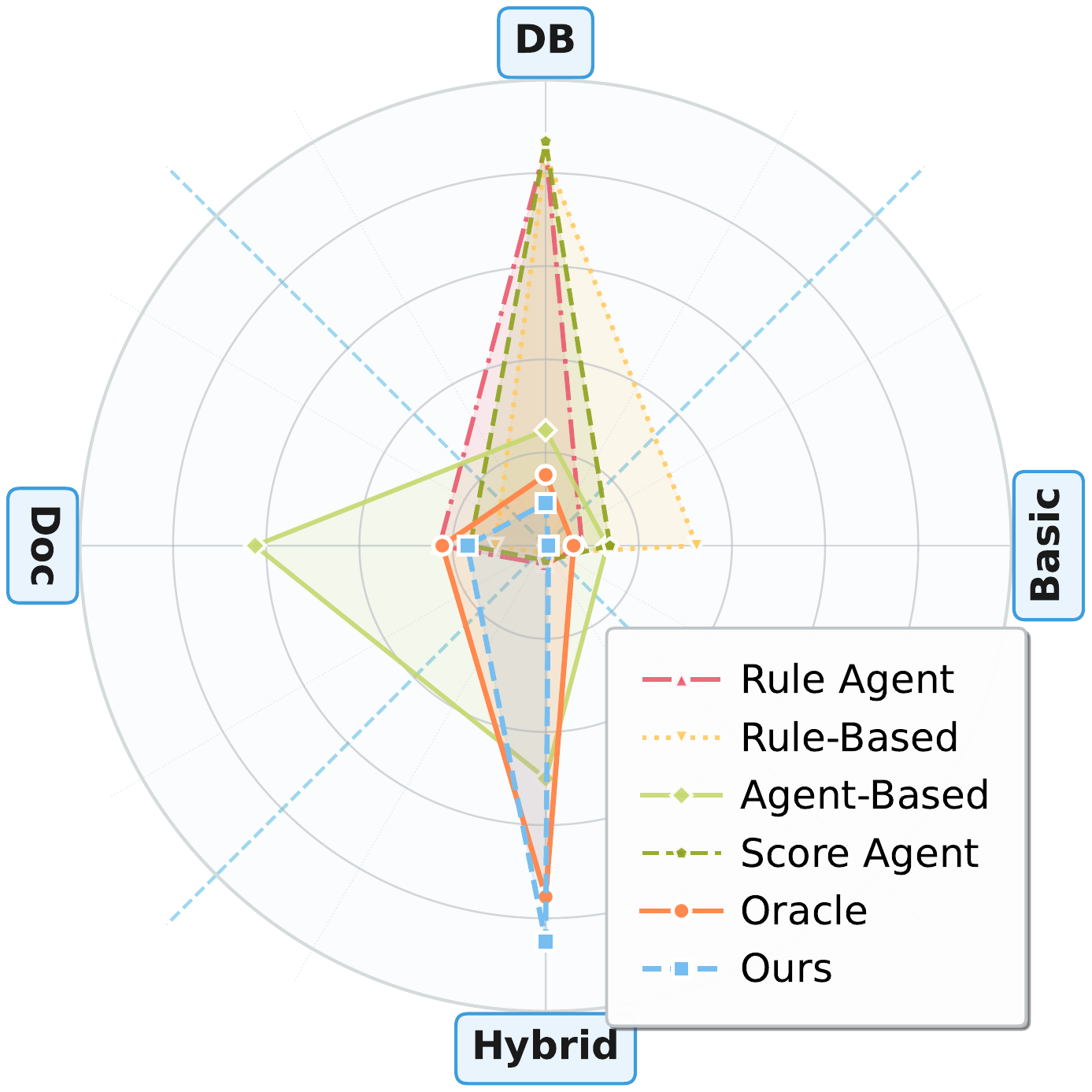}
  }
    \subfloat[\centering 
    \label{fig:PSR_finqa}
    \small{Path Selection on FINQA}]{
    \includegraphics[width=0.22\textwidth]{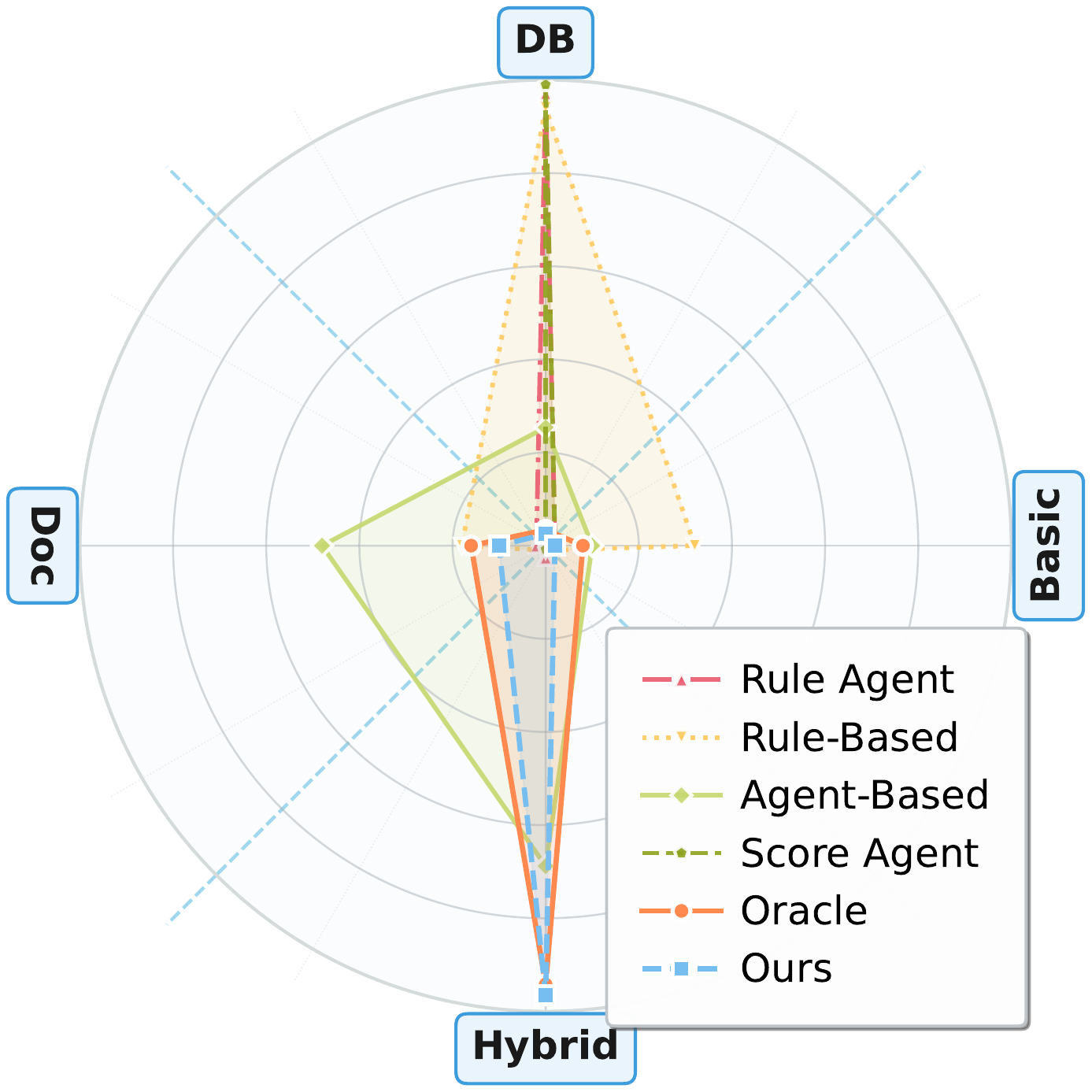}
  }
\caption{\small{Investigation of Path Selection. 
For better visualization, the Basic and Doc results are proportionally scaled up while preserving their relative ratios.}}
\label{fig:Path}
\end{figure}

\subsection{Investigation of Path Selection}

To better understand how our method allocates queries across different augmentation strategies, we further analyze the distribution of selected paths. 
Figure~\ref{fig:Path} reports the path selection profiles on TATQA and FinQA under the Qwen2.5 backbone. 
Each radar plot shows the proportion of queries routed to the DB, Doc, Hybrid, and Basic paths, comparing our method with several baselines and an oracle strategy. 
The oracle represents an upper bound where each query is assigned to the augmentation path that yields the correct answer. 
We observe that baseline strategies exhibit skewed or inconsistent allocations: rule-based and rule-agent methods often over-assign queries to DB, while agent-based and score-agent approaches tend to misallocate queries to suboptimal paths. 
In contrast, our method produces a distribution that closely matches the oracle, with a suitable use of DB, Doc, and Hybrid paths according to the query. 
This demonstrates that our routing mechanism is able to approximate oracle-level path decisions, effectively leveraging the complementary strengths of heterogeneous knowledge sources.

\section{Conclusion} In this work, we investigated the complementary roles of relational databases and unstructured documents in retrieval-augmented question answering. Our analyses showed that these sources excel on different query types: databases are most effective for fact-centric and numerical questions, while documents better handle open-ended or descriptive queries. We further demonstrated that naïve hybrid augmentation is not a solution, as it increases token overhead and latency without consistent accuracy gains, underscoring the necessity of routing. To this end, we introduced a rule-driven routing framework that grounds path selection in explicit, interpretable rules, continuously refines them through feedback with a rule-making expert agent, and accelerates inference with a path-level meta-cache. Extensive experiments on three QA benchmarks confirmed that our method consistently outperforms static and learned routing baselines, achieving higher accuracy while controlling cost. Overall, this study highlights the importance of principled routing across structured and unstructured knowledge sources, paving the way for more accurate, efficient, and interpretable RAG systems.

\newpage
\bibliographystyle{ACM-Reference-Format}
\bibliography{ref}
\newpage
\appendix

\section{Case Study}\label{case_study}

\begin{figure}[ht]
\centering
\includegraphics[width = 0.5\textwidth]{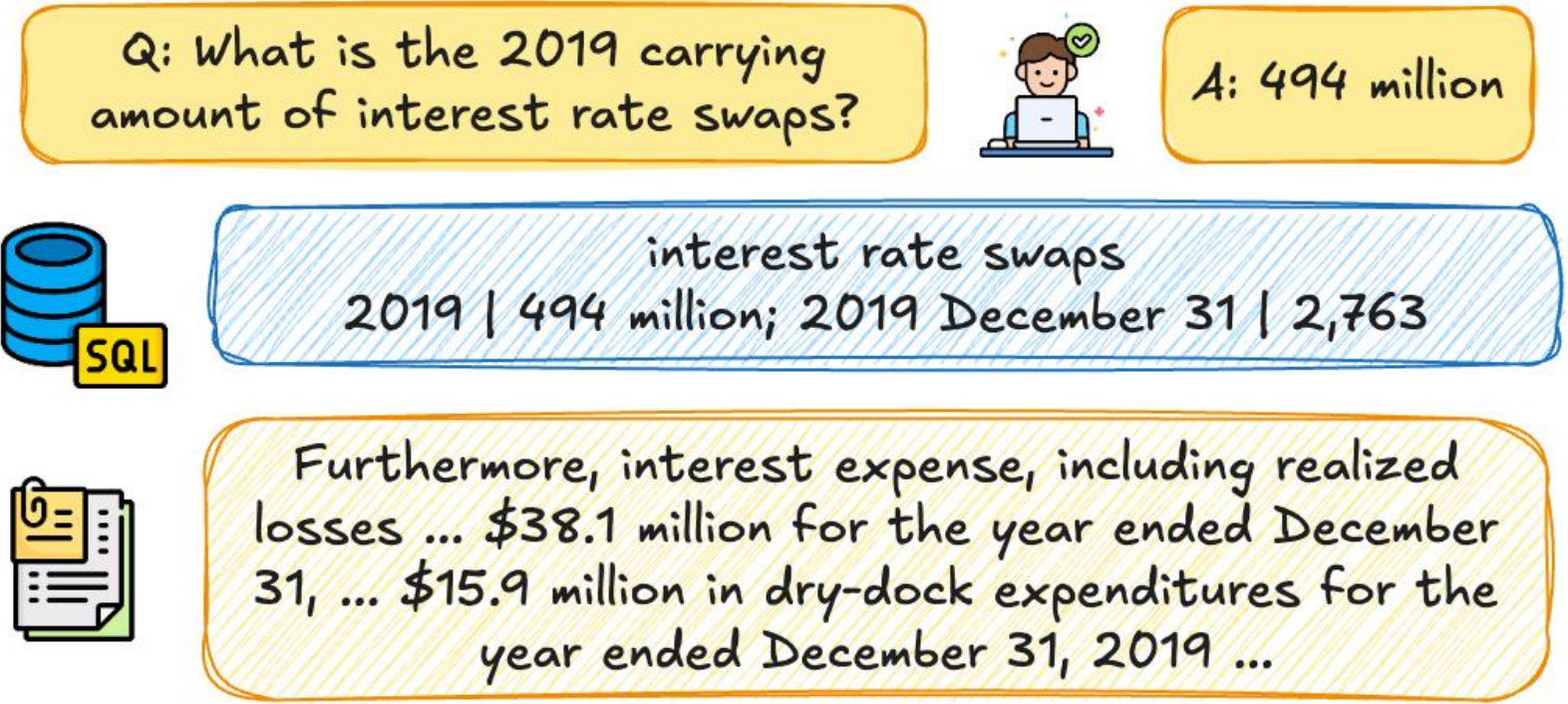}
\caption{\small{Case Study}}
\label{fig:case}
\end{figure}

To further illustrate the necessity of routing, we present a concrete example drawn from the TATQA dataset (Figure~\ref{fig:case}). The query is \textit{"What is the 2019 carrying amount of interest rate swaps?"}, and the ground-truth answer is \texttt{494 million}.  
In this case, the correct answer can only be obtained from the database path. The DB-only model outputs \texttt{494 million}, which exactly matches the ground truth. By contrast, the Doc-only model fails, since the retrieved passage only discusses related financial concepts such as interest expenses and cross-currency swaps but never provides the carrying amount. More importantly, the Hybrid model, which merges database and document evidence, is misled by repeated mentions of \textit{December 31, 2019} in the document. These distractions bias the model toward the wrong numerical entry from the database, producing the incorrect answer \texttt{2,763}. This demonstrates that naïve combination of structured and unstructured sources can actually harm performance, as irrelevant contextual details from documents dilute the reliability of precise database facts.  
Our rule-driven routing framework avoids this failure by correctly assigning the query to the DB path. According to Rule~1 ("If a question requests numbers, percentages, years, or calculations, then FACT\_ONLY path +3"), the system recognizes the numeric nature of the question and prioritizes the database. This not only delivers the correct answer but also ensures efficiency: the DB-only path consumes far fewer tokens than the Hybrid path, which incurs additional computational cost without providing accuracy gains.

\begin{figure*}[ht]
\centering
\includegraphics[width = 0.85\textwidth]{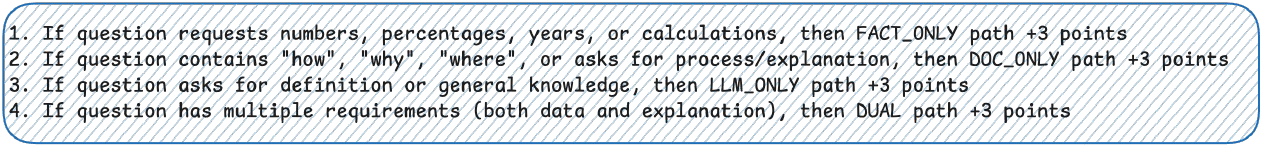}
\caption{\small{Initial Rule}}
\label{fig:rule_ori}
\end{figure*}

\begin{figure*}[ht]
\centering
\includegraphics[width = 0.85\textwidth]{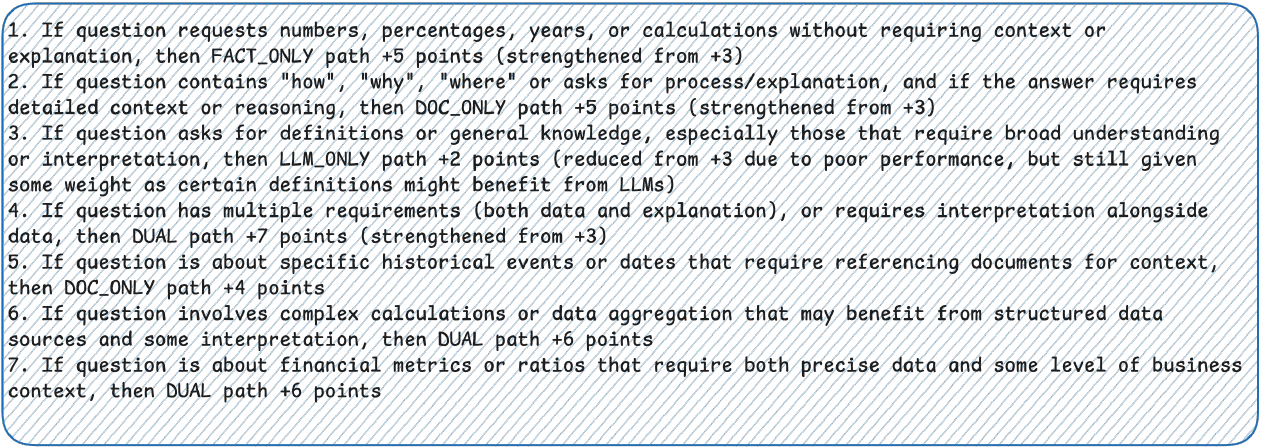}
\caption{\small{TATQA Updated Rule}}
\label{fig:rule_tatqa}
\end{figure*}

\begin{figure*}[ht]
\centering
\includegraphics[width = 0.85\textwidth]{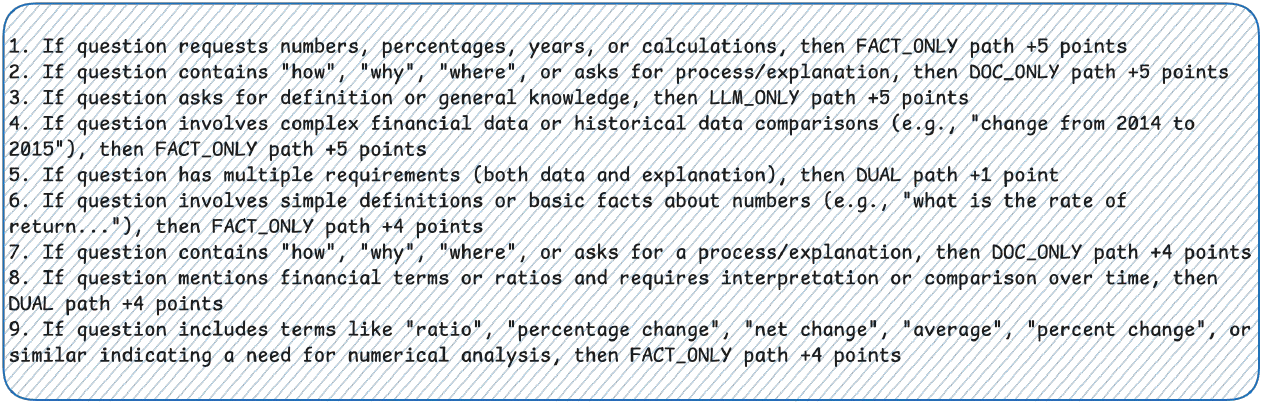}
\caption{\small{FINQA Updated Rule}}
\label{fig:rule_finqa}
\end{figure*}

\begin{figure*}[ht]
\centering
\includegraphics[width = 0.85\textwidth]{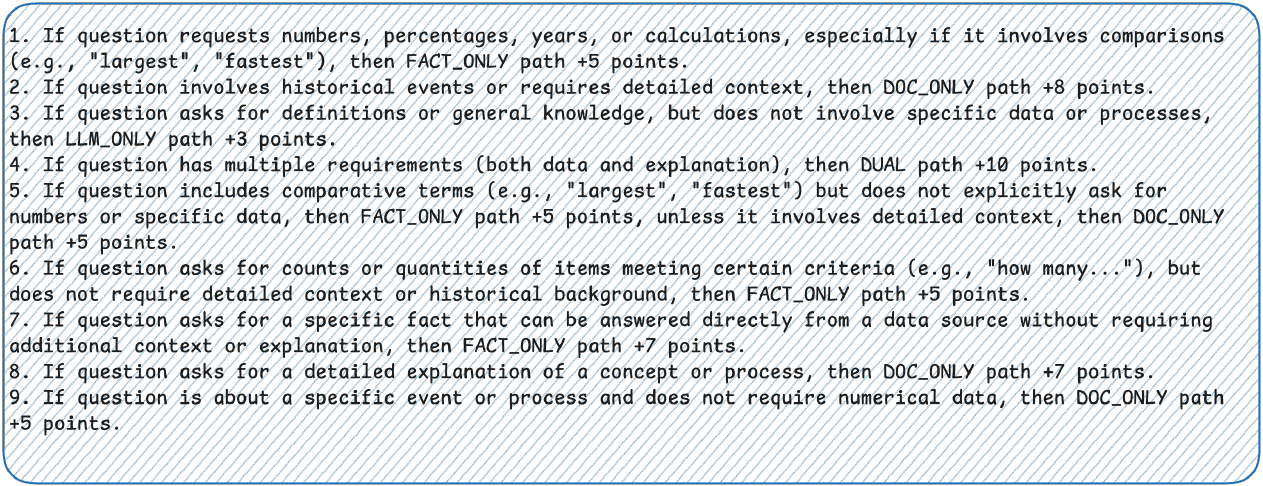}
\caption{\small{WIKIQA Updated Rule}}
\label{fig:rule_wikiqa}
\end{figure*}

\begin{figure*}[ht]
\centering
\includegraphics[width = 0.85\textwidth]{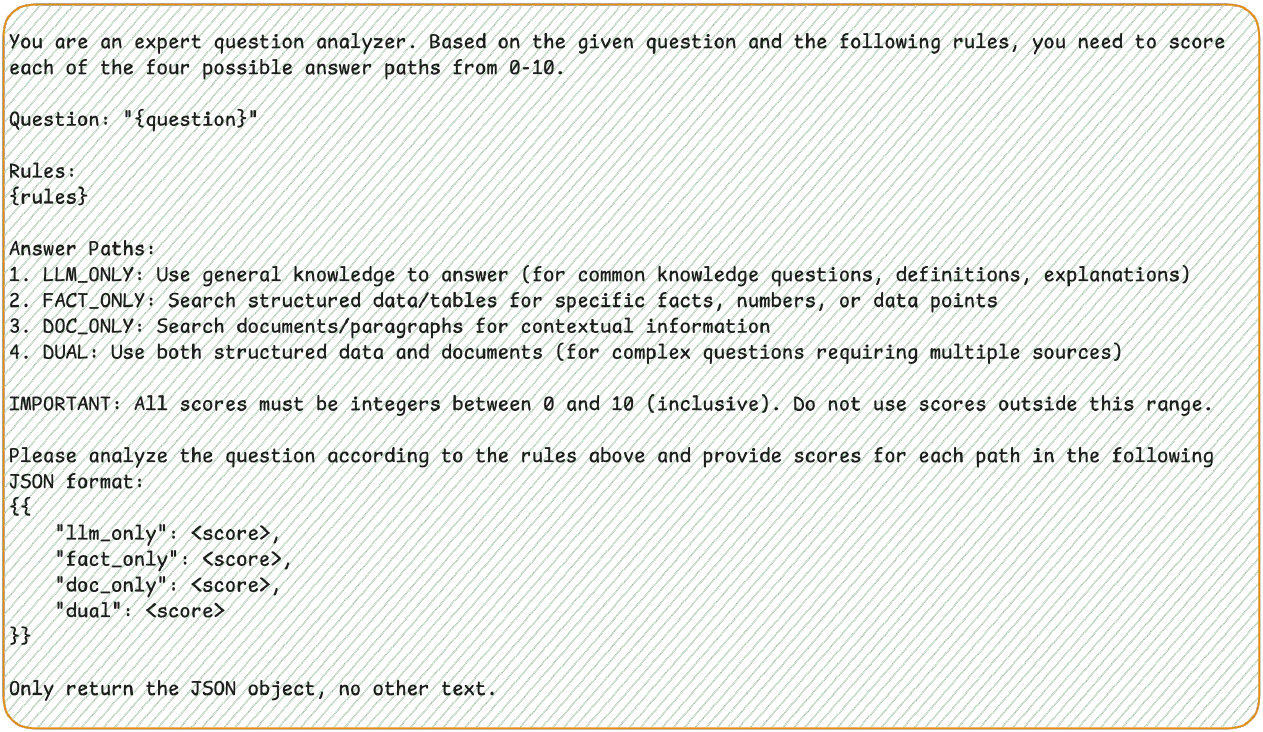}
\caption{\small{Prompt for Rule-Driven Routing Agent}}
\label{fig:prompt_sco}
\end{figure*}

\begin{figure*}[ht]
\centering
\includegraphics[width = 0.85\textwidth]{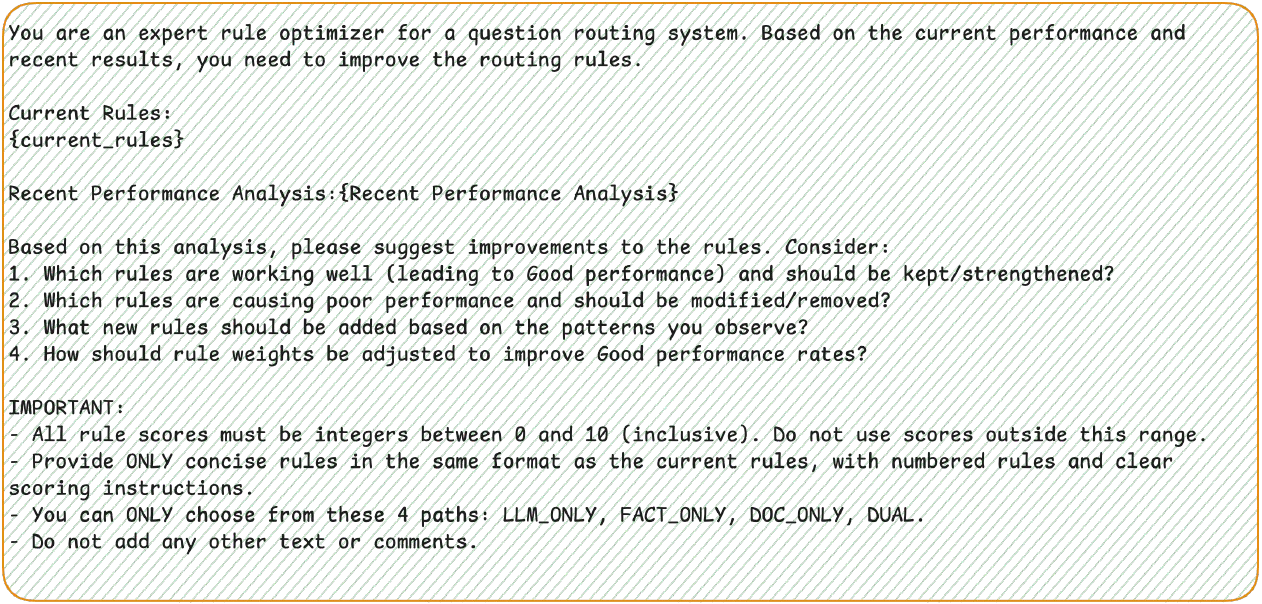}
\caption{\small{Prompt for Rule-Driven Routing Agent}}
\label{fig:prompt_rule}
\end{figure*}

\section{Rule Updates} \label{appendix:rule}
Figures~\ref{fig:rule_ori}--\ref{fig:rule_wikiqa} present the evolution of routing rules used in our framework. 
Figure~\ref{fig:rule_ori} shows the initial hand-crafted rules, while Figures~\ref{fig:rule_tatqa}, \ref{fig:rule_finqa}, and \ref{fig:rule_wikiqa} illustrate the updated rules learned on TATQA, FinQA, and WikiQA, respectively. 
These results are obtained using the Qwen2.5 backbone and demonstrate how the rule-making expert agent incrementally refines the rule set based on query–answer feedback. 
The updates capture dataset-specific patterns (e.g., numerical queries in TATQA, financial reasoning in FinQA, and open-domain descriptions in WikiQA), confirming that our rule-driven routing framework can adaptively specialize rules to different domains.

\section{Prompts} \label{appendix:prompt}
Figures~\ref{fig:prompt_sco} and \ref{fig:prompt_rule} provide the full prompt templates used in our routing framework. 
Figure~\ref{fig:prompt_sco} shows the prompt for the score-based routing agent, where the LLM is instructed to evaluate candidate augmentation paths and return numerical scores. 
Figure~\ref{fig:prompt_rule} shows the prompt for the rule-driven routing agent, where the LLM is guided to interpret rules and apply them to the current query. 
Together, these prompts define the interaction between the LLM and our routing components, ensuring reproducibility and clarity of the experimental setup.
\end{document}